     \documentclass{article}%
     \usepackage{fullpage}
     \usepackage{amsmath,amssymb, amsfonts}
     \usepackage{booktabs}
     \usepackage{array, graphicx}
     \usepackage{hyperref}
     \usepackage{tikz}
     \usetikzlibrary{bayesnet,fit}
     \renewcommand{\edge}[2]{\foreach \source in {#1}{\foreach \target in {#2}{\draw[->] (\source) -- (\target);}}}
     \renewcommand{\plate}[4][]{\node[draw, rounded corners, fit=#3, inner sep=4pt, #1] (#2) {};\node[anchor=south east, inner sep=2pt] at (#2.south east) {#4};}
     \tikzset{HSCMarrow/.style={->, thick, shorten <=2pt, shorten >=2pt}}
     
     \graphicspath{{Fig/}}
     \onecolumn
     \usepackage{bm}
     \usepackage{caption}

     \newcommand{\Exp}[1]{\mathbb{E}\left[#1\right]}
     \newcommand{\EExp}[2]{\mathbb{E}_{#1}\left[#2\right]}
     \newcommand{\rmdo}{\mathrm{do}}
     \newcommand{\g}{\,\vert\,}
     \newcommand{\pr}{\mathrm{p}}
     \newcommand{\di}{\mathrm{d}}

     \usetikzlibrary{arrows.meta, positioning, shapes.geometric}
     \usepackage{slashed}

     \newcommand{\V}{\mathcal{V}}

     \newcommand{\G}{{\mathcal{G}}}

     \newcommand{\U}{\mathcal{U}}
     \newcommand{\Do}{\mathrm{do}}

     \usepackage{natbib}
     
\begin{document}
     
     % \journaltitle{Journal of the Royal Statistical Society Series C}
     % \DOI{}
     % \copyrightyear{2026}
     % \pubyear{2026}
     % \vol{}
     % \issue{}
     % \access{}
     % \appnotes{Methodology}
     
     % \firstpage{1}
     
     \title{Scalable and Versatile Identification for Hierarchical Structural Causal Models: A New Look at Project STAR}

    \author{Janis Aiad, Aghiles Drali, Aymen El Ouadrhiri, Anass Ettahiri,\\Yasser Oufqir, Simon Patry,\\ David Cortes, Marianne Clausel, Emilie Devijver}
     
     % \address[1]{\orgname{Ecole Polytechnique}, \orgaddress{\country{france}}}
     % \address[2]{\orgname{AI-vidence}, \orgaddress{\country{france}}}
     % \address[3]{\orgdiv{CNRS, Laboratoire d'Informatique de Grenoble}, \orgname{Universite Grenoble Alpes}, \orgaddress{\country{france}}}
     % \address[4]{\orgdiv{CRAN, Simul Research Group}, \orgname{Université de Lorraine}, \orgaddress{\country{france}}}
     
     % \corresp[$\ast$]{Corresponding author. \href{mailto:emilie.devijver@univ-grenoble-alpes.fr}{emilie.devijver@univ-grenoble-alpes.fr}}
     
     % \received{Date}{0}{Year}
     % \revised{Date}{0}{Year}
     % \accepted{Date}{0}{Year}

    % \keywords{Hierarchical Structural Causal Models, Project STAR, Class-size effects, Automatic Identification, Do-Calculus, Scalable Estimation}
     
    % \boxedtext{Key Messages}{
     %\parbox{0.94\textwidth}{\raggedright%
     %We provide a statistical pipeline for automatic HSCM identification and scalable estimation from fitted local probability models. Applied to the project STAR, the pipeline turns class heterogeneity into a causal estimand and gives a new interpretation of class-size effects beyond non-causal hierarchical Bayesian/OLS baselines.\\\phantom{a}}}
     
     \maketitle

      \abstract{%
     The STAR (Student-Teacher Achievement Ratio) experiment (1985, Tennessee, USA) is a landmark hierarchical dataset designed to assess the impact of class size on student outcomes, with observations nested within classes. 
     To encode class-level interventions in such hierarchical settings, we develop a complete, scalable, open-source pipeline for Hierarchical Structural Causal Models (HSCM) that bridges symbolic identification and practical estimation.
     Our approach integrates graph transformations, \texttt{pyAgrum}'s do-calculus for automatic identification of causal effects, adaptation of symbolic expression into closed-form HSCM formulas, and numerical estimation from fitted local probability models.
     A key innovation is our adapted Abstract Syntax Tree (AST), which decomposes \texttt{pyAgrum}’s identified formulas into independent density, expectation, and marginalization tasks, enabling parallel and scalable computation. 
     We validate the pipeline on canonical HSCM motifs and benchmark scenarios with known ground truth, then apply it to STAR kindergarten mathematics outcomes.
     The results show that flat baselines (ignoring hierarchy) recover associations but fail to encode class-level interventions, and that symbolic identification alone is not enough for practical Hierarchical Structural Causal inference; scalable estimation and numerical stability checks are central parts of the scientific object.
     The code, including the HSCM implementation and STAR replication scripts, is available \href{https://github.com/ai-vidence/hierarchicalcausalmodels}{here}. 
     }
     % \makeatletter
     % \def\@evenhead{\hbox to \textwidth{\sffamily\fontsize{8bp}{10bp}\selectfont
     % {{\fontseries{m}\fontsize{9bp}{10bp}\selectfont\thepage}}\hspace*{4pt}\raisebox{-1.5pt}{\rule{.3pt}{8pt}}\hspace*{4pt}{\strut\@journaltitle, \@pubyear}\hfill}}
     % \def\@oddhead{\hbox to \textwidth{\hfill\sffamily\fontsize{8bp}{10bp}\selectfont
     % {{\strut\@journaltitle, \@pubyear}}\hspace*{4pt}\raisebox{-1.5pt}{\rule{.3pt}{8pt}}\hspace*{4pt}{{\fontseries{m}\fontsize{9bp}{10bp}\selectfont\thepage}}}}
     % \makeatother
     
     %% =============================================================================
     \section{Introduction}
     \label{sec:intro}
     %% =============================================================================
     
     Project STAR is a landmark randomized study in education, conducted in Tennessee in 1985, initiated by \citet{Krueger1999} and later analyzed for long-run impacts of class assignment \citep{Krueger2001,Chetty2011}. STAR assigned {in a fully randomized way} students within schools to one of three class types: small classes, regular classes, or regular classes with a teacher’s aide. Outcomes included standardized test scores in mathematics and reading, measured at the student level. While the canonical econometric analysis relies on student-level regressions with school fixed effects, the treatment mechanism is inherently hierarchical: class type is shared by all students in the same class, and class composition (such as gender balance, socioeconomic status, or peer environment) can influence both outcomes and the interpretation of the treatment effect. Crucially, the intervention (class size) is assigned at the class level, yet its effects are measured at the student level. The intervention target is therefore naturally modelled as a%is therefore a 
     distribution of sub-unit treatments within a unit—%not a 
     and not as a binary treatment assigned independently to each %row 
     observation (here the students) in a flat dataset.
     
     Classical causal inference provides a framework for identifying causal estimands from observed data \citep{Pearl2009,Imbens2015}, available for example in the library \texttt{pyAgrum} \citep{pyagrum}. It relies on the ID algorithm or the adjustment formulae when all variables are observed, and allow to rewrite a causal effect from observational distribution. However, most models assume a flat set of random variables, which fails to capture the nested structure of many real-world datasets, and, in particular, for our primary case study, Project STAR. In Project STAR, a flat model cannot distinguish between a class-level intervention (e.g., reducing class size for all students in a class) and a student-level intervention (e.g., moving a single student to a smaller class). If the analyst aggregates students into class means, within-class heterogeneity (e.g., peer effects, teacher-student interactions) is lost. If the analyst pools all students, unit-level latent variables (e.g., class composition) induce unobserved dependence and confounding.

     \citet{weinstein2023HSCM} introduced the formal framework for Hierarchical Structural Causal Models (HSCM) to model hierarchidal data (e.g., sutdents nested in classes). They propose their graphical representation, identification via {an extended version of }do-calculus and estimation. 
     Unlike flat models, HSCM explicitly encode the hierarchical structure, enabling analysts to model interventions at the {unit-level, instead of restricting to individual-level interventions only} %correct level 
     (that is, in our running example class-level instead student-level). %This framework thus addresses the core limitation of flat models: their inability to represent unit-level interventions. 
     Once an HSCM estimand is identified, it is usually not a single regression parameter but a complex formula combining unit-level probability models, response curves, and averages over {auxiliary variables, especially introduced to handle HSCM.} 
     Without {an adapted} structured representation, {such a hierarchical estimation procedure} becomes a monolithic formula that is hard to audit and hard to scale.
     Existing work enables analysts to model hierarchical interventions 
     and propose estimators of the causal effect, but stops short of providing a complete pipeline from raw nested data to numerical estimates for general hierarchical causal graph. The notebook associated to \citet{weinstein2023HSCM} is illustrating the method on a particular example, and the \texttt{y0} library \citep{y02025} recently added symbolic HSCM support via a domain-specific language, enabling { the exploration of alternative scenarios, paving the way to counterfactual analysis.}%counterfactual transport 
     
   %  {\color{blue}La aussi c'est pas clair, on a pas vu le contrefactuel, et on a pas relie contrefactuel et transport. On voit pas où il y a du calcul symbolique tant qu'on a pas parlé de do calculus, je simplifie aussi}. 
     
     However, these tools %focus on symbolic derivation and 
     leave critical operational steps unresolved: %adapting symbolic outputs into closed-form formulas, 
     fitting local probability models, scaling computations, and validating results on real-world datasets like those provided by Project STAR.
     This paper addresses these challenges by operationalizing HSCMs. Starting from a hierarchical graph, our implementation collapses the graph, augments it with outcome-distribution nodes when needed, and marginalizes {the variables that will not be involved in the causal estimand}. %irrelevant variables 
    % {\color{blue}De quel point de vue les variables sont irrelevant? Elles interviennent pas dans le do-calculus? Du coup elles sont pas irrelevant?}. 
     It then calls \texttt{pyAgrum} \citep{pyagrum} for symbolic do-calculus identification, adapts the returned expression into closed-form HSCM formulas, and evaluates the resulting functional using probability and response models fitted from the nested data.
     The novelty of our approach relies in the use of the Abstract Syntax Tree (AST) to record the formal estimand, drive numerical evaluation, and exposes which density factors were actually fitted.
     %The identified expression is not merely a display object, it is converted into an abstract syntax tree (AST) that simultaneously records the formal estimand, drives numerical evaluation, and exposes which density factors were actually fitted.
     {The use of an AST provides major benefits in terms of estimation scalability. By decomposing the identified formula into explicit computational units, its hierarchical representation enables each node—such as a conditional density, expectation, or summation—to be estimated independently. The overall formula is then evaluated by combining many small, interpretable estimators rather than relying on a single opaque global model.}
     
     %The AST representation enables scalable estimation: each node (conditional density, expectation, or summation) is an explicit unit of work. Thus, the same identified formula can be evaluated by fitting many small, interpretable estimators rather than a single opaque global model. 
     We validate the pipeline on synthetic benchmarks and real-world data, focusing on STAR’s kindergarten mathematics outcomes. 
     {Our results demonstrate that reliable hierarchical structural causal inference requires the interplay between hierarchical modeling and symbolic methods. In particular, ignoring the hierarchical structure (e.g., by using flat models) may recover associations but fails to capture unit-level interventions. Conversely, symbolic identification alone is insufficient; scalable estimation and numerical stability checks are necessary for reliable inference.}

     Our main contributions are:
     \begin{itemize}
     \item a full, scalable and operational pipeline for estimating causal effects in hierarchical data, available as open-source Python code at \href{https://github.com/AI-vidence/hierarchicalcausalmodels}{\texttt{AI-vidence/hierarchicalcausalmodels}};
     \item an illustration of the method in the analysis of STAR's data, illustrating how hierachical models outperform flat models. 
     \end{itemize}

     The remainder of the paper is organized as follows.
     Section~\ref{sec:problem} introduces standard notions of SCMs and HSCM, Section~\ref{sec:method} describes the symbolic-to-numeric method, Section~\ref{sec:experiments} validates the implementation on controlled settings, Section~\ref{sec:star} gives the Project STAR analysis, and Section~\ref{sec:conclusion} discusses limitations and future work.
     The code, including the HSCM implementation and STAR replication scripts, is available in \href{https://github.com/AI-vidence/hierarchicalcausalmodels}{this repository}.\footnote{https://github.com/AI-vidence/hierarchicalcausalmodels}

     %% =============================================================================
     \section{Preliminaries: causal inference and hierarchical models}
     \label{sec:problem}
     %% =============================================================================

     \subsection{Structural Causal Models and interventions.} 
     A \emph{Structural Causal Model} (SCM, \citet{Pearl2009}) is defined as $\mathcal{M}:=\left< \bm{{U}}, \bm{{V}}, {P}({\bm{{U}}}) \right>$ where $\bm{U}$ is a set of exogenous variables taking values in $\mathcal{U}$, $\bm{V} = \{V_1,\ldots, V_d\}$ is a set of endogenous variables taking values in $\mathcal{V} = \mathcal{V}_1\times \ldots \times \mathcal{V}_d$, $\mathcal{F} = \{f_1,\ldots, f_d\}$ is a set of functions such that for each $1\leq j \leq d$, $f_j:({pa}_j, {u}_j)\in (\mathcal{V}_{\mathrm{Pa}(V_j)} \times \mathcal{U}_j ) \mapsto v_j \in \mathcal{V}_j$ and ${P}(\bm{U})$ is a probability distribution over mutually independent exogenous variables $\bm{U}$, with strictly positiveness on $\mathcal{U}$. ${\mathrm{Pa}}(V_j)$ is the set of parents of $V_j$ and $pa_j$ its realization. We assume that the SCM $\mathcal{M}$ induces a \emph{directed acyclic graph} (DAG) $\mathcal{G}=(\bm{V},\mathcal{E})$ with an edge from $V_i$ to $V_j$ whenever $V_i\!\in\!\mathrm{Pa}(V_j)$. The ancestor set $\mathrm{An}(V_j)$ contains all nodes with a directed path to $V_j$; $\mathrm{De}(V_j)$ denotes descendants. 
     
     In a given SCM \(\mathcal{M}=\langle \U,\V,\mathcal{F},P(\U)\rangle\), let \(\mathbf{V}=(V_1,\dots,V_n) \) be a subset of \(\V\) and \(\mathbf v\) be a valuation of \(\mathcal V\). A hard intervention \(\Do(\mathbf{V}=\mathbf{v})\) replaces each structural function \(f_i\) for \(V_i\in\mathbf{V}\) by the constant assignment \(V_i:=v_i\), which in the causal diagram \(\G_{\mathcal{M}}\) corresponds to removing all incoming edges into each \(V_i\). 
     A soft intervention assigns possibly
     non-constant functions to variables without adding new causal connections.
     
     \subsection{Hierarchical Structural Causal models.}
      
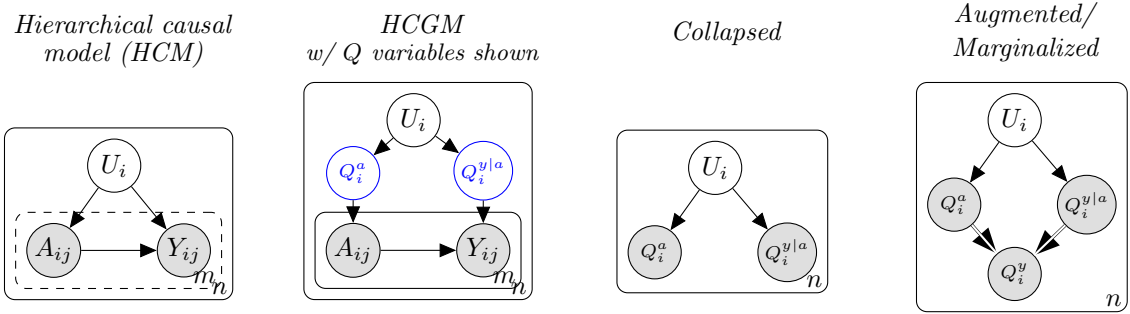
\begin{figure}[t!]
\centering

% Premier diagramme (Confounder)
\begin{minipage}{0.24\textwidth}
\centering
\begin{tikzpicture}
\node[obs] (y) {$Y_{ij}$};
\node[obs, left=1cm of y] (a) {$A_{ij}$};
\node[latent, above=0.4cm of a, xshift=0.8cm] (u) {$U_i$};

\edge {a,u} {y};
\edge {u} {a};

\plate[dashed] {in} {(a)(y)} {$m$};
\plate {out} {(in)(u)} {$n$};

\node at (-0.8,3) {\textit{Hierarchical causal}};
\node at (-0.8,2.6) {\textit{model (HCM)}};
\draw[white, opacity=0] (current bounding box.north west) rectangle (current bounding box.south east);
\end{tikzpicture}
\label{fig:hcm_confounder}
\end{minipage}%
%
% Deuxième diagramme (HCGM)
\begin{minipage}{0.24\textwidth}
\centering
\begin{tikzpicture}
\node[obs] (y) {$Y_{ij}$};
\node[obs, left=1cm of y] (a) {$A_{ij}$};
\node[latent, above=1 of a, xshift=0.8cm] (u) {$U_i$};
\node[latent, above=.3 of a, color=blue, fill=blue!0] (qa) {$\color{blue} \scriptstyle Q^{a}_i$};
\node[latent, above=.3 of y, color=blue, fill=blue!0] (qya) {$\color{blue} \scriptstyle Q^{y \mid a}_i$};

\edge {u} {qya};
\edge {qya} {y};
\edge {u} {qa};
\edge {qa} {a};
\edge {a} {y};

\plate {ay} {(a)(y)} {$m$};
\plate {ayu} {(ay)(u)} {$n$};

\node at (-0.8,3) {\textit{HCGM}};
\node at (-0.8,2.6) {\small \textit{w/ $Q$ variables shown}};
\draw[white, opacity=0] (current bounding box.north west) rectangle (current bounding box.south east);
\end{tikzpicture}
\label{fig:hcm_intro_hpgm}
\end{minipage}%
%
% Troisième diagramme (Collapsed)
\begin{minipage}{0.24\textwidth}
\centering
\begin{tikzpicture}
\node[obs] (y) {$\scriptstyle Q_i^{y \mid a}$};
\node[obs, left=1cm of y] (a) {$\scriptstyle Q_{i}^{a}$};
\node[latent, above=0.4cm of a, xshift=0.8cm] (u) {$U_i$};

\edge {u} {y};
\edge {u} {a};

\plate {ayu} {(a)(y)(u)} {$n$};

\node at (-0.8,2.9) {\textit{Collapsed}};
\draw[white, opacity=0] (current bounding box.north west) rectangle (current bounding box.south east);
\end{tikzpicture}
\label{fig:collapse_intro}
\end{minipage}%
%
% Quatrième diagramme (Augmented/Marginalized)
\begin{minipage}{0.24\textwidth}
\centering
\begin{tikzpicture}
\node[obs] (y) {$\scriptstyle Q_i^{y \mid a}$};
\node[obs, left=1cm of y] (a) {$\scriptstyle Q_{i}^{a}$};
\node[latent, above=0.4cm of a, xshift=0.8cm] (u) {$U_i$};
\node[obs, below=.2cm of a, xshift=0.8cm] (qy) {$\scriptstyle Q_i^{y}$};

\edge {u} {y};
\edge {u} {a};
\path (a) edge [very thick, ->] (qy);
\path (a) edge [color=white, thick] (qy);
\path (y) edge [very thick, ->] (qy);
\path (y) edge [color=white, thick] (qy);

\plate {ayu} {(a)(y)(u)(qy)} {$n$};

\node at (-0.8,2.5) {\textit{Augmented/}};
\node at (-0.8,2.05) {\textit{Marginalized}};
\draw[white, opacity=0] (current bounding box.north west) rectangle (current bounding box.south east);
\end{tikzpicture}
\label{fig:augment_intro}
\end{minipage}

\caption{\textbf{An example of hierarchical causal model and its transformation.} (a) an HSCM   with $n$ units, each of which containing $m$ subunits, (b) HCGM with explicit HCM's latent $Q$ variables, (c) HCM's matching collapsed model and (d) its augmented/marginalized version. We apply do-calculus to this final model to perform identification in the original HCM.} \label{ex:fig1}
\end{figure}

     In an HSCM, endogenous variables are partitioned into \emph{unit-level} (e.g., classes) and \emph{subunit-level} (e.g., students) variables, with distributions defined at both levels. 
     {The idea of HSCM is to introduce a new concept of intervention at the unit level alongside the classical intervention at the individual level.}
     
     {To explain it properly, let us illustrate this on an example depicted in Figure~\ref{fig:pipeline}. In this running example, we are given $n$ units (for e.g. classes), each containing $m$ subunits (for e.g. $m$ students in each class). The treatment variable for each individual $j$ of the unit $i$ is denoted $A_{ij}$ (for e.g. tutoring hours for student $j$ in class $i$) whereas the corresponding outcome is denoted $Y_{ij}$ (for example mathematic grade of student $j$ in class $i$ at the end of the year). We assume that we have a common counfounder at the unit level, denoted $U_i$ for each unit $i$, that may describe for example the size of the class. }
     {Classically, interventions on variables $A_{ij}$ can be performed at the individual level. HCSM paves the way to a new kind of interventions, namely unit level interventions, by introducing auxiliary variables, the so called $Q$-variables. 
     They allow to transform the Structural Causal Model into a Bayesian hierarchical model by adding a new level of hierarchy related to interventional distributions (respectively conditional distributions with respect to interventions). For e.g., in our running example, one introduces two auxiliary variables $Q^{(a)}$ and $Q^{(y|a)}$ which are respectively a distribution over interventional distributions on $A$ and a distribution over conditional distributions of $Y$ conditionally to $A$.  
     The HSCM framework consists in turning the causal model  into a generative model. We explicit it for the example in  Figure~\ref{fig:pipeline}:
     \begin{itemize}
     \item We draw i.i.d. counfouders $U_i$ modelling the common covariates for each unit $i$ (in our example for each class)
     \item Thereafter we generate the individual treatments $A_{i,j}$ in the following way:
     \begin{itemize}
     \item Conditionally to $U_i$, we draw a distribution $Q_i^{(a)}$ on interventional distribution common for each unit depending on a parameter $a$. In the running example corresponding to the STAR project modelling, for each class, we draw a distribution on interventional distributions of the number of hours of tutoring for each student.
     \item We draw from the distribution $Q_i^{(a)}$ the treatment variables for each individual $j$ in unit $i$. In our running example, for each student we draw a distribution of the number of hours of tutoring for each student $j$ in unit $i$.
     \end{itemize}
     \item Now we generate the individual outcomes $Y_{ij}$ in the following way:
     \begin{itemize}
     \item Conditionally to $U_i$, we draw a distribution on conditional interventional distribution $Q_i^{(y|a)}$ common for each unit $i$
     \item We draw from $Q_i^{(y|a)}$ the outcome $Y_{ij}$ of each individual $j$ of each unit $i$.
     \end{itemize}
     \end{itemize}

     %We focus our analysis to hard interventions on unit variables and sort interventions on subunit variables, which may condition on other subunit variables. 

     The $Q$-variables are then  summarizing subunit-level mechanisms (e.g., treatment assignment, outcomes) at the unit level, enabling causal reasoning in hierarchical settings. 
     %Let $i=1,\ldots,n$ index units and $j=1,\ldots,m_i$ index sub-units, for each sub-unit variable $L_{ij}$ with sub-unit parents $\pa_S(L)_{ij}$, we consider a unit-level $Q$-node
     %\begin{equation}
     %Q_i^{L\g \pa_S(L)}
     %=
     %\pr\!\left(L_{ij}\g \pa_S(L)_{ij}\right).
     % \label{eq:qdef}
     %\end{equation}
     This aggregation allows us to define causal targets that operate on distributions of subunit-level variables rather than individual observations.
     Our primary causal target is the contrast between expected unit-level outcomes under two different distributions of subunit-level treatments: 
     \begin{equation}
     \tau(q_0\rightarrow q_1) = \mathbb{E}[Y \mid \Do(Q^{(a)} = q_1)] - \mathbb{E}[Y \mid \Do(Q^{(a)} = q_0)],
     \label{eq:problem-target}
     \end{equation}
     which compares the expected unit-level outcome distribution under two different treatment distributions. Other target may be considered, and similar analysis can be derived, based on the identification of each term. We restrict here (without loss of generality) to the average treatment effect. }

     In general, to apply standard causal identification tools (e.g., do-calculus) to HSCMs, \citet{weinstein2023HSCM} propose a three-step graphical transformation that converts the hierarchical model into an equivalent flat model:
     
     \begin{itemize} 
     \item \emph{Step 1: collapsing.} Replace every sub-unit mechanism by its $Q$-node  and lift all edges to a flat directed acyclic graph over unit-level variables and $Q$-variables. %Produce a flat model containing only the unit level variables $U_i$, $Q^{(a)}_i$ and $Q^{(y|a)}$.
     
     \item \emph{Step 2: augmenting.} Augment the collapsed model with the auxiliary variable $Q^{(y)}$ modelling the distributions of outcomes.
     
    \item  \emph{Step 3: marginalizing.} Project out variables that are not part of the identification problem.
     \end{itemize}
    
    %  marginalization projects out variables that are not part of the identification problem.
      
     %To transform the hierarchical model, we develop three types of graphical steps: collapsing, augmenting, and marginalizing. When we collapse, we promote the $Q$ variables to endogenous variables, and remove the subunit endogenous variables, forming a flat causal model; when we augment, we add new unit-level endogenous variables; and when we marginalize, we remove some of the unit-level endogenous variables. Finally, with the flat model in hand, we apply the do-calculus to determine whether and how identification is possible.
     Then, standard do-calculus can be applied to determine whether a causal target like \(\tau(q_0\rightarrow q_1)\) is identifiable from the observed data. This approach leverages the power of graphical methods while accounting for the hierarchical structure of the data.

     \section{Causal effect estimation in Hierarchical Structural Causal Models}
     \label{sec:method}
     %% =============================================================================
     
     \subsection{Overview of the approach}
     
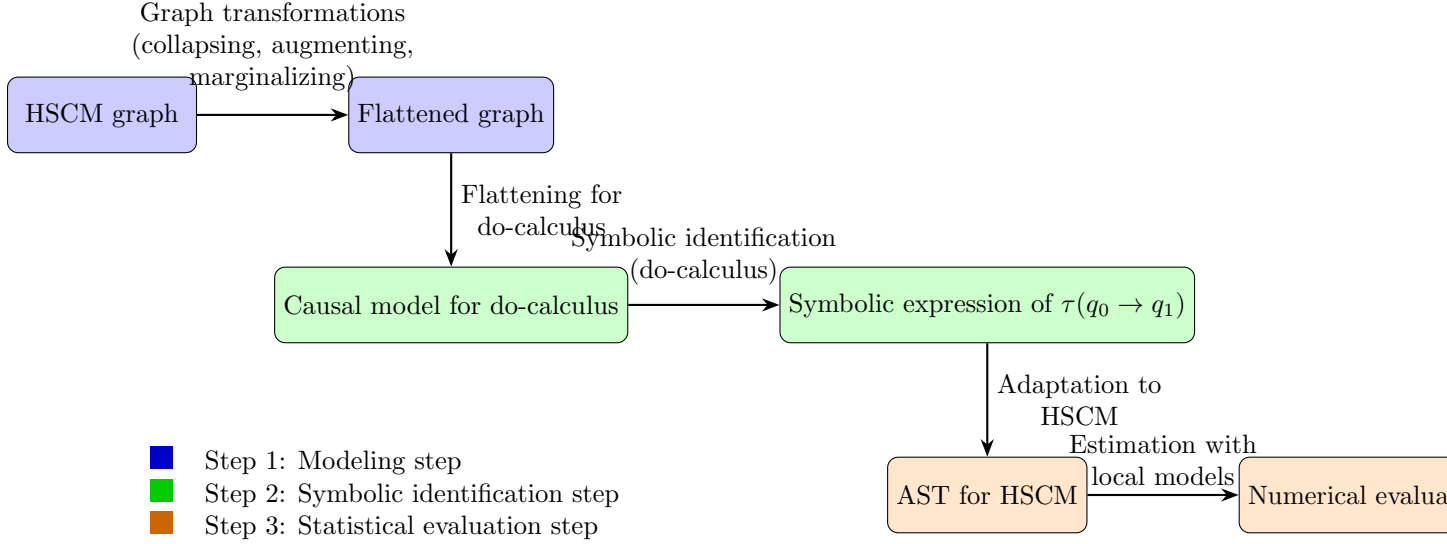
\begin{figure}[t]
     \centering
     \begin{tikzpicture}[
     node distance=1.5cm and 2cm,
     box/.style={draw, rectangle, rounded corners, minimum width=2.5cm, minimum height=1cm, align=center},
     tool/.style={draw, cylinder, shape border rotate=90, aspect=0.2, minimum width=2cm, minimum height=1.2cm, align=center},
     process/.style={draw, diamond, minimum width=2cm, minimum height=1.2cm, align=center},
     arrow/.style={-Stealth, thick},
     blue/.style={fill=blue!20},
     green/.style={fill=green!20},
     orange/.style={fill=orange!20}
     ]
     
     % Nodes
     \node[box, blue] (hscm) {HSCM graph};
     \node[box, blue, right=of hscm] (collapsed) {Flattened graph};
     \node[box, green, below=of collapsed] (pyagrum_model) {Causal model for do-calculus};
     \node[box, green, right=of pyagrum_model] (pyagrum_expr) {Symbolic expression of $\tau(q_0\rightarrow q_1)$};
     \node[box, orange, below=of pyagrum_expr] (ast) {AST for HSCM};
     \node[box, orange, right=of ast] (evaluator) {Numerical evaluator};
     
     % Arrows with labels
     \draw[arrow] (hscm) -- node[above=5pt, align=center] {Graph transformations \\ (collapsing, augmenting, \\ marginalizing)} (collapsed);
     \draw[arrow] (collapsed) -- node[right, align=center] {Flattening for \\ do-calculus} (pyagrum_model);
     \draw[arrow] (pyagrum_model) -- node[above=4pt, align=center] {Symbolic identification \\ (do-calculus)} (pyagrum_expr);
     \draw[arrow] (pyagrum_expr) -- node[right, align=center] {Adaptation to \\ HSCM} (ast);
     \draw[arrow] (ast) -- node[above, align=center] {Estimation with \\ local models} (evaluator);
     
     % Legend
     \node[left=3.2cm of ast] {
     \begin{tabular}{ll}
     \textcolor{blue!80!black}{\rule{0.3cm}{0.3cm}} & Step 1: Modeling step \\
     \textcolor{green!80!black}{\rule{0.3cm}{0.3cm}} & Step 2: Symbolic identification step \\
     \textcolor{orange!80!black}{\rule{0.3cm}{0.3cm}} & Step 3: Statistical evaluation step \\
     \end{tabular}
     };
     
     \end{tikzpicture}
     \caption{Pipeline for causal identification in Hierarchical Structural Causal Models (HSCMs).}
     \label{fig:pipeline}
     \end{figure}

     Our contribution is a complete and scalable pipeline that bridges the gap between hierarchical causal models (HSCMs) and practical estimation, by combining existing tools in a novel and principled way. While each component (graphical transformations, symbolic identification, numerical evaluation) exists in isolation, their integration into a single, automated workflow for hierarchical data is, to our knowledge, new. This pipeline enables analysts to move seamlessly from a hierarchical graph and nested data to a numerical estimate of causal contrasts like $ \tau(q_0\rightarrow q_1)$ introduced in Eq.~\ref{eq:problem-target}.  
     The pipeline consists of three stages, illustrated in Figure~\ref{fig:pipeline}:
     \begin{enumerate}
     \item[Step 1] \textbf{Modeling step}: From the HSCM, we construct a flat causal graph  using the $Q$-variable trick, collapsing, augmenting, and marginalizing steps of \citet{weinstein2023HSCM}. This step replaces subunit-level variables with \(Q\)-variables (e.g., class-level treatment distributions) and ensures the graph is compatible with standard do-calculus.
     
     \item[Step 2] \textbf{Symbolic identification step}: The flattened graph is passed to \texttt{pyAgrum} \citep{pyagrum}, which performs do-calculus to derive a symbolic expression for the causal target (e.g., \(\tau(q_0\rightarrow q_1)\)). This expression is a functional of observed distributions, but it is initially agnostic to the hierarchical structure. Our contribution here is to translate this output back into the HSCM framework mapping each term in the expression to a hierarchical mechanism %(e.g., \(P(Q^{y \mid a,g} \mid S)\)).
     %{\color{blue}J'ai commenté car j'ai pas compris qui c'est qui S?}
     
     \item[Step 3] \textbf{Statistical evaluation}: The translated expression is decomposed into a set of independent estimation tasks (e.g., estimation of conditional densities, expectations, or marginalizations), each of which can be fitted locally to the data. %For example, class-level proportions \(Q^{(a)}a\) may be estimated via logistic regression, while outcome distributions \(Q^{(y|a)}\) may use Gaussian or mixture models. 
     %
    % {\color{blue}J'ai laissé $Y$ car pour le moment on est pas specialisé sur le projet STAR}
     %
     The modularity of this step, done by Abstract Syntax Tree (AST), allows for parallel estimation across units, factors, or Monte Carlo samples, making the pipeline scalable to large hierarchical datasets.
     \end{enumerate}
     
     This separation between symbolic and numerical layers is key: the symbolic layer determines \emph{what} needs to be estimated (based on the graph and intervention), while the numerical layer handles \emph{how} to estimate each component (using local models).  %, which we discuss in Section~\ref{sec:diagnostics}.

     % The implementation is organized as a symbolic-numeric backend rather than as a collection of graph-specific estimators.
     % The analyst provides a hierarchical graph whose variables are typed by level, together with nested arrays whose rows are units and whose columns are sub-units.
     % from that input, the backend follows a single path from the graphical specification to a numerical contrast:

     Step 1, the modeling step, has been introduced in generality in \citet{weinstein2023HSCM} and summarized in our preliminaries (Section~\ref{sec:problem}). 
     
     Step 2 consists in identifying a causal effect $\mathbb{E}(Y | \Do(Q^{(a)} = q))$, for a flatten graph: the routine implemented in \texttt{pyAgrum}, based on the ID algorithm, is diretly doing that. As we assume causal sufficiency, adjustment by the parent is one way for example to identify. 
    Remark that \texttt{pyAgrum} never sees the original nested table: it sees the collapsed HSCM graph and returns a symbolic expression for $\mathbb{E}[Y\g\Do(Q^a=q^a)]$ in terms of observed variables of that graph. Keeping this step separate from estimation is what allows the same identified expression to be evaluated later with Gaussian, mixture, categorical, non-parametric, or neural models for the required local probabilities and responses.

     Now we describe the last step of our procedure, the numerical evaluation. 
    Particularly, we describe in details the Abstract Syntax Tree (AST) structure in Step 3, which ensures that the pipeline is both \textbf{general} (applicable to any HSCM) and \textbf{efficient} (exploiting parallelism inherent in the hierarchical structure). Importantly, it also allows for diagnostics (e.g., checking the stability of the final contrast $\tau(q_0\rightarrow q_1)$ to perturbations in local models). Then, we illustrate how to estimate the formula on two examples, but the pipeline can consider any estimation method.

     \subsection{Step 3: Statistical evaluation of the causal hierarchical effects}
     %\subsection{Estimators: pros and cons}
     The general target is $\tau(q_0\rightarrow q_1)$ given in Equation \eqref{eq:problem-target}, seen as a difference between the two numerical evaluations, related to the interventions $Q^{(a)}=q_0$ and $Q^{(a)}=q_1$. We explain how to estimate one term: changing from $\rmdo(Q^a=1)$ to $\rmdo(Q^a=0)$ does not change the method, only the intervention value used. 
    
     \subsubsection{Abstract Syntax Tree for estimation}
     %\subsection{Synthetic Benchmarks}
     \label{app:synthetic-benchmarks}
     
     %Operationally, Equation~\eqref{eq:qdef} is treated as a rule for constructing a new node name and its parents.
     %for example, an outcome mechanism $M\leftarrow(A,E,G,L)$ becomes the node $Q^{m\g a,e,g,l}$.
     %The backend stores this node as a variable of the collapsed graph, but its numerical value is only fitted later from the sub-unit table.
     %Thus equation~\eqref{eq:qdef} has two roles: graphically, it replaces a sub-unit mechanism by a unit-level variable; numerically, it declares which conditional density estimator will be needed.
    
     %The final step consists of converting the flattened graph obtained from the HSCM transformations into the corresponding \texttt{pyAgrum} objects. We describe how the graph structure is encoded in a relevant way.
     %The graph passed to \texttt{pyAgrum} is obtained through the HSCM transformations.

The symbolic expression returned by \texttt{pyAgrum} is adapted into the Abstract Syntax Tree (AST) data type of \texttt{pyAgrum} and then written as the closed-form do-calculus formula used by the estimator.
The AST data type is a nested object representation of the \texttt{pyAgrum}-identified formula after translation into HSCM density factors.
Each node of the AST has a computational role.
A summation node stores the variables to marginalize; a product node stores the multiplicative factors; a conditional node stores one density key of the form $P(\mathrm{outcome}\g\mathrm{parents})$; and a leaf node is used for constants or fallback terms.
The do-calculus engine demo prints this adapted object directly.
The example below is not the STAR graph: it is a small synthetic collapsed-graph example using the same transformation logic as Figure~\ref{ex:fig1}, with two extra variables added only to make the AST traversal visible.
It contains an outer marginalization over $Q^{z\g a}$ and $W$, an inner marginalization over $Q^a$, and four fitted density factors:

     \begin{center}
     \setlength{\fboxsep}{7pt}
     \begin{tabular}{@{}c@{\hspace{0.035\textwidth}}c@{}}
     \textbf{Abstract Syntax Tree (AST)} & \textbf{Identified formula} \\[4pt]
     \fbox{%
     \begin{minipage}[t]{0.43\textwidth}
     \ttfamily\scriptsize
     sum on Qz\_a,W for\par
     \textbar{} *\par
     \textbar{} \textbar{} *\par
     \textbar{} \textbar{} \textbar{} P(W\textbar{}Qa)\par
     \textbar{} \textbar{} \textbar{} P(Y\textbar{}Qa,Qz\_a,W)\par
     \textbar{} \textbar{} sum on Qa for\par
     \textbar{} \textbar{} \textbar{} *\par
     \textbar{} \textbar{} \textbar{} \textbar{} P(Qz\_a\textbar{}Qa,W)\par
     \textbar{} \textbar{} \textbar{} \textbar{} joint P(Qa)
     \end{minipage}}%
     &
     \fbox{%
     \begin{minipage}[t]{0.40\textwidth}
     \vspace{0.35em}
     \centering
     \scriptsize
     \[
     \begin{aligned}
     &\sum_{Q^{z\g a},W}
     P(W\g Q^a)\,
     P(Y\g Q^a,Q^{z\g a},W)\\
     &\quad\times
     \left[
     \sum_{Q^a}
     P(Q^{z\g a}\g Q^a,W)\,
     P(Q^a)
     \right].
     \end{aligned}
     \]
     \vspace{0.35em}
     \end{minipage}}%
     \end{tabular}
     \end{center}
     %This is the object used by the evaluator.
    On the right, we give an causal estimand and on the left, is detailed the way it is encoded in the AST.
     The first line of the AST is the outer marginalization node over $Q^{z\g a}$ and $W$.
     The product node of the AST then has two children: a direct product of fitted conditionals, and an inner summation over $Q^a$.
     From this tree, we need to have access to the density keys $P(W\g Q^a)$, $P(Y\g Q^a,Q^{z\g a},W)$, $P(Q^{z\g a}\g Q^a,W)$, and $P(Q^a)$, that must be fitted or read from data.
     In larger formulas, this same mechanism may produce dozens of density keys and many repeated unit-level prediction problems.
     Those tasks share the same formal estimand but do not need to be fitted as a single global regression, which is why the framework can scale to large nested datasets without changing the causal definition of the target.

     \subsubsection{Estimation procedure}
     %In the implementation, it is evaluated by copying the identified AST twice.
     %The first copy fixes the intervention node at $Q^a=q_1$, the second fixes it at $Q^a=q_0$, and the ATE is the difference between the two numerical evaluations.
 The estimation step depends on the graph and the quantity to be estimated. The AST returns the conditional factors that must be estimated, and the numerical backend then selects an estimator according to the variable type and the available parent set. In the current implementation, discrete variables can be fitted with Bernoulli or categorical/multinomial estimators, continuous score variables with Gaussian or two-component Gaussian mixture estimators, and other supported families include Poisson, exponential, Gamma, Beta, lognormal, inverse-Gaussian, Student, Laplace, half-Cauchy, and non-parametric density estimators. We illustrate our estimation procedure over two classical causal patterns, but the same AST-to-estimator dispatch can be used for general HSCM formulas.

     {We stress the inherent scalability of our algorithm, which naturally derives from its modular implementation. Indeed, each conditional density factor can be estimated from its own parent set and then each $\widehat\mu_i$ can be fitted from the sub-units of unit $i$, before averaging.}
     
     \paragraph{Counfounder case}
  We come back to the confounder case used as the running example, where a latent unit variable $U_i$ affects both treatments $(A_{ij})$ and outcomes $(Y_{ij})$ of each individual $j$ of the unit $i$.
     The collapsed graph contains $Q_i^a$ and $Q_i^{y\g a}$ as unit-level objects, and the intervention is evaluated by averaging the per-unit response: it reduces to the familiar {Difference In Means estimator considered at the unit level}%unit-averaged estimator
     \begin{equation}
     \widehat \tau(q_0\rightarrow q_1)
     =
     \frac{1}{n}\sum_{i=1}^n
     \left[
     \widehat\mu_i(q_1)-\widehat\mu_i(q_0)
     \right],
     \qquad
     \widehat\mu_i(q)
     =
     \int y\, \widehat Q_i^{(y\g a)}(y\g q)\,\di y.
     \label{eq:plugin}
     \end{equation}
     %Here, the outer average is an explicit loop over units. The term $\widehat\mu_i(q)$ is produced by the estimator attached to the conditional $Q$-node $Q_i^{(y\g a)}$ for parametric families the integral is analytic or quadrature-based; for simulation-based components the backend draws Monte Carlo samples and averages the predicted outcomes.
     {To compute the term $\widehat\mu_i(q)$, we considered in our implementation two cases. For parametric families the integral is analytic or quadrature-based. In the non parametric setting, the backend draws Monte Carlo samples and averages the predicted outcomes.}
     %More complex graphs use the same evaluator, but the AST contains more density factors and more marginalization nodes.
    
     %Scalable estimation enters at exactly this point.
     %Each $\widehat\mu_i$ can be fitted from the sub-units of unit $i$, each conditional density factor can be estimated from its own parent set, and each branch of a sampled marginalization can be evaluated independently before the AST combines the results.
     %The computational structure therefore mirrors the statistical structure: many local estimators are assembled into one causal functional.
    % {\color{blue}In the sequel, we shall also consider a more general case where we have instrumental variables.}
     %The canonical motifs used in the experiments serve as validation cases for this machinery.
     %They verify that \texttt{pyAgrum} identification, HSCM formula adaptation, and numerical evaluation agree on structures whose behaviour is known in advance.
    % {\color{blue}Le coufounder c'est une redite on en a déjà parlé. Du coup je remets pas eux fois la même chose }
     %\begin{itemize} 
     %\item Let us consider the counfounder case where a latent unit variable $U_i$ affects both treatments $(A_{ij})$ and outcomes $(Y_{ij})$ of each individual $j$ of the unit $i$.
     %The collapsed graph contains $Q_i^a$ and $Q_i^{y\g a}$ as unit-level objects, and the intervention is evaluated by averaging the per-unit response:
     %\begin{equation}
     % \Exp{Y\g \rmdo(Q^{(a)}=q_\star)}
     %=
     %\EExp{i}{ \int y\, Q_i^{(y\g a)}(y\g q_\star)\,\di y
     %}.
     %\label{eq:confounder-canonical}
     %\end{equation}
     
     %\item 
     \paragraph{Instrumental variable}

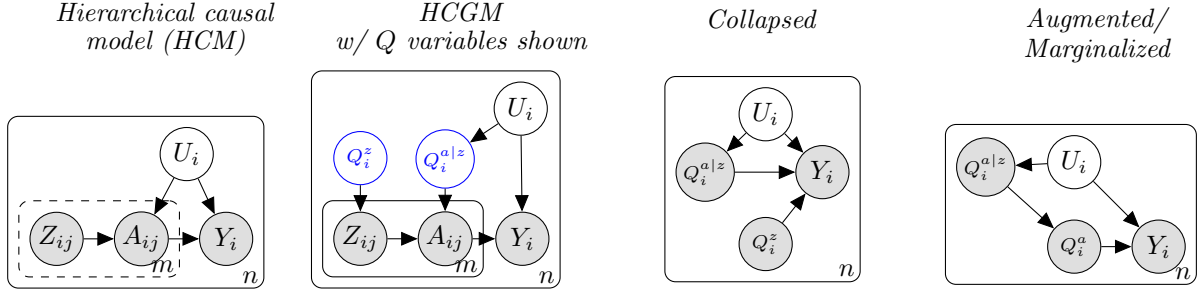
\begin{figure}[t!]
% Premier diagramme (Confounder)
\begin{minipage}{0.24\textwidth}
\centering
\begin{tikzpicture}

  \node[obs]                               (y) {$Y_{i}$};
  \node[obs, left=0.4cm of y] (a) {$A_{ij}$};
  \node[obs, left=0.4cm of a] (z) {$Z_{ij}$};
  \node[latent, above=0.4cm of a, xshift=0.6cm]  (u) {$U_i$};

  \edge {a,u} {y} ;
  \edge {u} {a} ;
  \edge {z} {a} ;

  \plate[dashed] {in} {(a)(z)} {$m$} ;
  \plate {out} {(in)(u)(y)} {$n$} ;

\draw[white, opacity=0] (current bounding box.north west) rectangle (current bounding box.south east);
\node at (-0.8,3) {\textit{Hierarchical causal}};
\node at (-0.8,2.6) {\textit{model (HCM)}};

\end{tikzpicture}
\end{minipage}
\begin{minipage}{0.24\textwidth}
\centering
\begin{tikzpicture}

  \node[obs] (y) {$Y_{i}$};
  \node[obs, left=0.3cm of y] (a) {$A_{ij}$};
  \node[obs, left=.4cm of a] (z) {$Z_{ij}$};
  \node[latent, above=1cm of a, xshift=1cm]  (u) {$U_i$};
  \node[latent, above=.35 of z, color=blue, fill=blue!0] (qz) {$\color{blue} \scriptstyle Q^{z}_i$};
  \node[latent, above=.3 of a, color=blue, fill=blue!0] (qaz) {$\color{blue} \scriptstyle Q^{a \mid z}_i$};

  \edge {u} {y} ;
  \edge{u} {qaz} ;
  \edge {qaz} {a} ;
  \edge {qz} {z} ;
  \edge {z} {a} ;
  \edge {a} {y} ;

  \plate {in} {(a)(z)} {$m$} ;
  \plate {out} {(in)(u)(y)(qaz)(qz)} {$n$} ;
\draw[white, opacity=0] (current bounding box.north west) rectangle (current bounding box.south east);

\node at (-0.8,3) {\textit{HCGM}};
\node at (-0.8,2.6) {\textit{w/ $Q$ variables shown}};

\end{tikzpicture}
\end{minipage}
\begin{minipage}{0.24\textwidth}
\centering
\begin{tikzpicture}

  \node[obs]                               (y) {$Y_{i}$};
  \node[obs, left=0.8cm of y] (qaz) {$\scriptstyle Q^{a\mid z}_i$};
  \node[obs, below=.2cm of qaz, xshift=0.8cm] (qz) {$\scriptstyle Q^{z}_i$};
  \node[latent, above=1cm of qz]  (u) {$U_i$};

  \edge {u} {qaz, y} ;
  \edge {qaz,qz} {y} ;

  \plate {out} {(qz)(qaz)(u)(y)} {$n$} ;
\draw[white, opacity=0] (current bounding box.north west) rectangle (current bounding box.south east);

\node at (-0.8,2) {\textit{Collapsed}};
\end{tikzpicture}
\end{minipage}
\begin{minipage}{0.24\textwidth}
\centering
\begin{tikzpicture}

  \node[obs]                               (y) {$Y_{i}$};
  \node[obs, left=1.5cm of qya] (qaz) {$\scriptstyle Q^{a\mid z}_i$};
  \node[obs, left=.4cm of y] (qa) {$\scriptstyle Q^{a}_i$};
  \node[latent, above=.4cm of qa]  (u) {$U_i$};

  \edge {u} {qaz, y} ;
  \edge {qa} {y} ;
  \edge {qaz} {qa} ;

  \plate {out} {(qaz)(u)(y)} {$n$} ;
\draw[white, opacity=0] (current bounding box.north west) rectangle (current bounding box.south east);

\node at (-0.8,3) {\textit{Augmented/}};
\node at (-0.8,2.6) {\textit{Marginalized}};
\end{tikzpicture}
\end{minipage}

\caption{\textbf{An example of hierarchical causal model and its transformation.} (a) an HSCM   with $n$ units, each of which containing $m$ subunits, (b) HCGM with explicit HCM's latent $Q$ variables, (c) HCM's matching collapsed model and (d) its augmented/marginalized version. We apply do-calculus to this final model to perform identification in the original HCM.} 
 \label{fig:hcm_instrument}
\end{figure}
    
     The instrumental variable case corresponds to the situation, where additionally to a common counfounder $U_i$ at the unit level to the treatment variables $A_{ij}$ and outcomes $Y_{ij}$, we add instrumental variables $Z_{ij}$ at the individual level, which are parents of the $A_{ij}$'s.
   For the instrumental motif, we get the following expression, including a reweighting term analogous to inverse probability weighting:
     \begin{equation}
     \Exp{Y\g \rmdo(A\sim q_\star^{(a)})}
     =
     \EExp{Q^{(a\g z)},Q^{(z)}}{
     \frac{q_\star^a(a)}{Q^{(a\g z)}((a\g z))}\,Y
     },
     \label{eq:iv-canonical}
     \end{equation}
     provided the positivity conditions are met.
     In the backend, the denominator $Q^{(a\g z)}(a\g z)$ is a fitted density node.
     If this estimator is missing or has no support at the sampled value, the calculation becomes unstable.
     The ratio $q_\star^{(a)}(a)/Q^{(a\g z)}(a\g z)$ is the likelihood-ratio weight between the target intervention law and the fitted treatment mechanism, exactly the quantity that appears in importance sampling and in Horvitz--Thompson/inverse-probability estimators \citep{RubinsteinKroese2016,HorvitzThompson1952,RobinsHernanBrumback2000}.
     This is why the implementation reports both the symbolic formula and factor-level diagnostics.
     %\end{itemize}
    
     \paragraph{Practical considerations}
     In practice, an estimation formula may contain high-variance conditional density factors.
     This can happen if the outcome have several parents, leading to the introduction of several $Q$-variables related to each parent of the outcome. For example in the running example of the STAR project, we may consider that the math grade distribution $Y$ depends jointly on treatment $A$, ethnicity $E$, gender $G$, and lunch status $L$.% such as $Q^{(y\g a,e,g,l)}$, where the outcome distribution $Y$ depends jointly in our running example on treatment $a$, ethnicity $e$, gender $g$, and lunch status $l$.
     
     The implementation records such cases and, in the normalized-factor runs studied below, rescales unstable multiplicative factors before evaluation. Operationally this is a controlled truncation of the raw product scale: the graph structure is retained, but the unstable factor is normalized so that it cannot overwhelm the rest of the functional.
     If the identified functional is $\mathfrak f$, the evaluated functional becomes a numerically stabilized object $\widetilde{\mathfrak f}$ with the same graph structure but normalized factor weights.
     This is not a harmless implementation detail: normalizing $P(Z\g W)$ changes the weighting induced by that factor and therefore changes the numerical functional being evaluated.
     We therefore report both the nominal identified formula and the factor normalization actually used in computation.

     \subsection{Computational cost}
     
     The computational efficiency of our pipeline stems from its modular design, which separates symbolic identification from numerical evaluation. This allows us to exploit parallelism at multiple levels (units, factors, and Monte Carlo samples), making the approach scalable to large hierarchical datasets. In Table \ref{tab:complexity}, we summarize the computational cost of each step. 
     
     The global complexity of the pipeline is dominated by the numerical evaluation step, and scales as \(O\left((nmd + BK)/P\right)\) under ideal parallelization. This reflects the fact that the symbolic steps (graph transformation and identification) are typically negligible compared to the numerical tasks, which involve fitting local models and Monte Carlo evaluation.
     \begin{table}[t]
     \centering
     \small
     \caption{Algorithmic cost and parallelization structure of the pipeline. The effective cost assumes ideal parallelization across $P$ workers. We denote $n$ is the number of units, $m$ is the average number of sub-units per unit, $d$ is the maximum parent count, $v$ and $e$ are the numbers of transformed graph nodes and edges, $K$ is the number of density factors in the identified formula, $B$ is the number of Monte Carlo samples or quadrature points.}
     \label{tab:complexity}
     \begin{tabular}{p{0.14\textwidth}p{0.25\textwidth}p{0.29\textwidth}p{0.25\textwidth}}
     \toprule
     Step & Sequential cost & Parallelization structure & Effective cost with $P$ workers \\
     \midrule
     Collapse/augmentation & $O(e+v)$ graph operations & graph-level (negligible) & $O(e+v)$ \\
     Build \texttt{pyAgrum} model & $O(e+v)$ (copy plus identification search) & symbolic backend & $O(e+v)$ \\
     AST translation & $O(K)$ expression traversal & factor-level traversal & $O(K)$ \\
     Per-unit $Q$ fitting & $O(nmd)$ for simple parametric factors & independent accross units and factors & $\sim O(nmd/P)$ + scheduling overhead \\
     Monte Carlo evaluation & $O(BK)$ per intervention & Independent across samples, units, and branches & $\sim O(BK/P)$ when branches are balanced \\
     \bottomrule
     \end{tabular}
     \end{table}

     %% =============================================================================
     \section{Experimental results over synthetic benchmarks}
     \label{sec:experiments}
     %% =============================================================================
     
%     \subsection{Canonical HSCM motifs}
The experimental validation is done with three canonical motifs: confounding, confounding with interference/frontdoor structure, and an instrumental-variable design.
These motifs are validation cases rather than restrictions of the pipeline: their ground truth is known, so they check that the graph transformation, identification, AST evaluation, and numerical estimators agree on controlled hierarchical structures.

\begin{figure}[t]
\centering
\begin{minipage}[t]{0.36\textwidth}
\centering
\vspace{0pt}
\includegraphics[width=\linewidth]{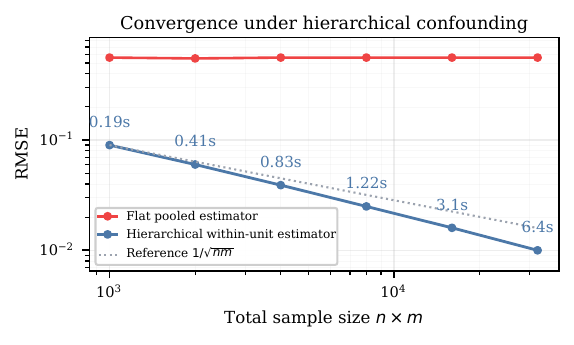}
\captionof{figure}{Convergence diagnostic under hierarchical confounding. The flat pooled estimator does not approach the intervention target because the unit-level latent cause is ignored. The hierarchical within-unit estimator removes the unit effect and follows the expected $1/\sqrt{nm}$ decay.}
\label{fig:synthetic-convergence}
\end{minipage}\hfill
\begin{minipage}[t]{0.60\textwidth}
\centering
\vspace{0pt}
\includegraphics[width=\linewidth]{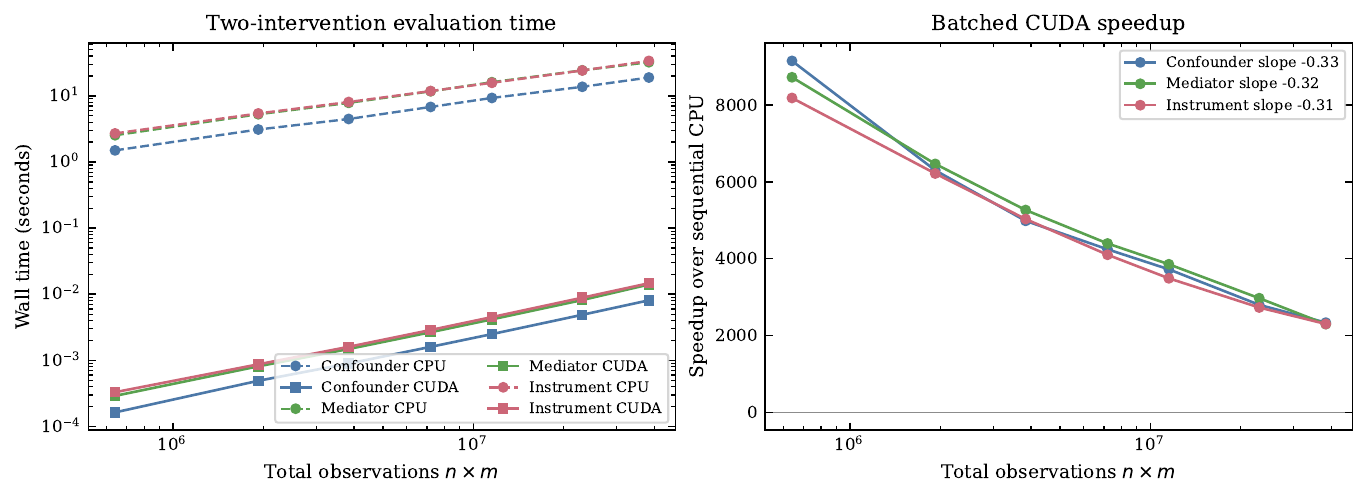}
\captionof{figure}{GPU speedup on synthetic HSCM evaluations. The left panel reports wall-clock time for sequential CPU evaluation and batched CUDA evaluation. The right panel reports speedup over sequential CPU; legend labels include the fitted log--log slope.}
\label{fig:gpu-speedup-cuda}
\end{minipage}
\end{figure}

The convergence diagnostic of the estimator in Figure~\ref{fig:synthetic-convergence} reports the confounder motif; the CUDA benchmark in Figure~\ref{fig:gpu-speedup-cuda} uses the same three motifs at larger synthetic sizes to isolate batching speed.
For a GPU implementation, the relevant parallelism is not a number of Python workers but a feasible tensor batch size, which depends on both $n$ and $m$.
We therefore treat GPU throughput as a hardware-specific benchmark rather than as a fixed worker-count comparison.
 The synthetic motifs are generated from known HSCMs, so they are used as controlled checks of the causal contrast rather than as real-data results.
     
     The confounder experiment, in Figure \ref{fig:synthetic-convergence}, also varies the number of units.
     With $m=50$ sub-units per unit, the estimate converges toward the true ATE as $n$ increases from 10 to 200, while pooled OLS remains biased because it mixes unit-level latent heterogeneity with the treatment effect.
     This behaviour is important for interpretation: the HSCM estimator is recovering the unit-level intervention encoded by the graph, rather than only fitting an association in the pooled student table.
     It is also important computationally:
     for these motifs, the expensive work consists of repeated unit-level response evaluations, factor evaluations, and Monte Carlo or quadrature branches.
     These tasks are independent conditional on the identified AST, so the sequential work $T_1$ can ideally be reduced to $T_P\simeq T_1/P$ with $P$ CPU workers, up to scheduling overhead.
     The same structure is highly GPU-parallelizable: units, sub-units, intervention values, and Monte Carlo samples are stacked into tensor batches and evaluated simultaneously when the density estimators are implemented in a tensor backend such as PyTorch or JAX\@.
     
     %\subsection{Synthetic Benchmarks}
     
     On log--log axes, the fitted slope for speedup is approximately $-0.32$ on average across motifs. This slope summarizes the measured RTX A5000 regime. It means that GPU remains thousands of times faster in absolute time, but the speedup decreases over the measured range because of memory traffic or occupancy limits become visible.

     %% =============================================================================
     \section{Real data analysis: Project STAR}
     \label{sec:star}
     %% =============================================================================
     
     \begin{figure}[b]
     
     \centering
     \begin{minipage}[t]{0.34\textwidth}
     \centering
     \vspace{0pt}
     \scriptsize
     \setlength{\tabcolsep}{2pt}
     \begin{tabular}{lll}
     \toprule
     Symbol & Level & Definition \\
     \midrule
     $A$ & student & Small-class assignment \\

    $Y$ & student & Mathematics score \\
    $B$ & student & Reading score \\
     $G$ & student & Gender indicator \\
     $E$ & student & Ethnicity indicator \\
     $L$ & student & free-lunch status\\
     $S$ & class & School urbanicity \\
     $U$ & latent & Unobserved class heterogeneity \\
     \bottomrule
     \end{tabular}
     \captionof{table}{STAR variables used in the HSCM analysis.}
     \label{tab:variables}
     \vspace{0.8em}
     \includegraphics[width=\linewidth]{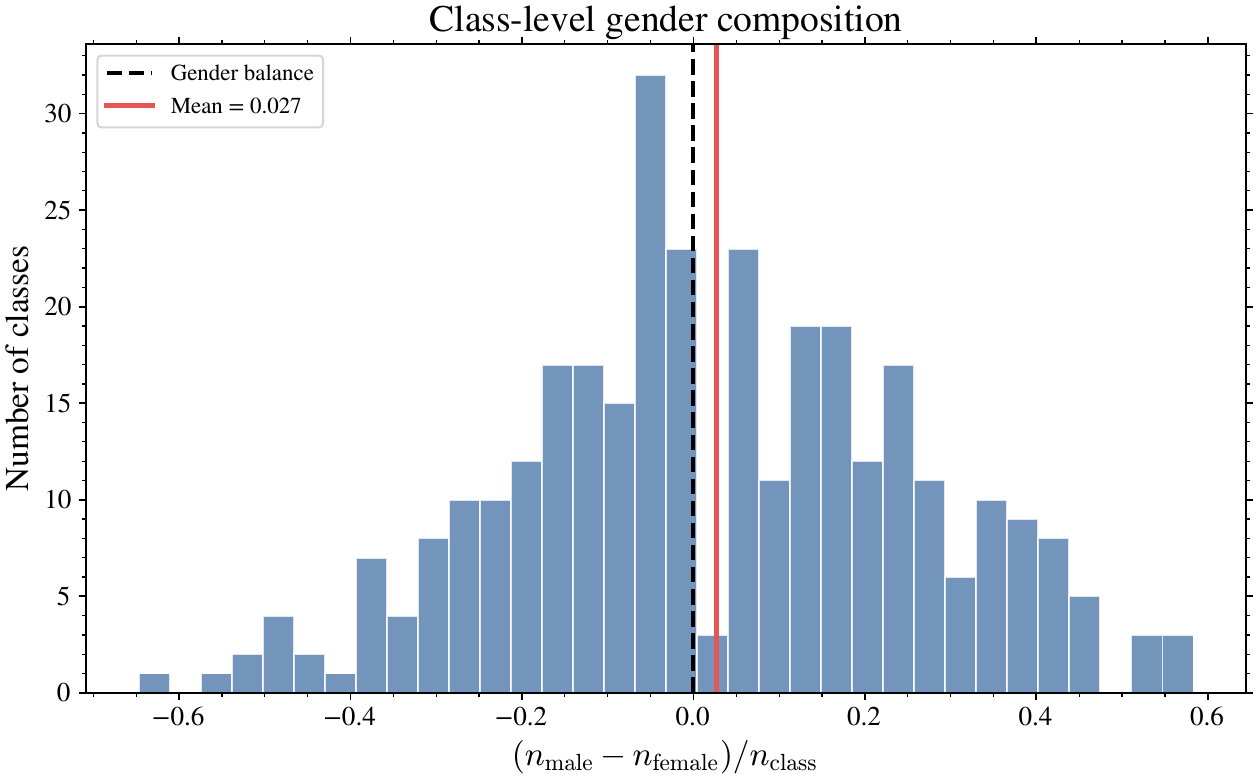}
     \end{minipage}\hfill
     \begin{minipage}[t]{0.62\textwidth}
     \centering
     \vspace{0pt}
     \includegraphics[width=\linewidth]{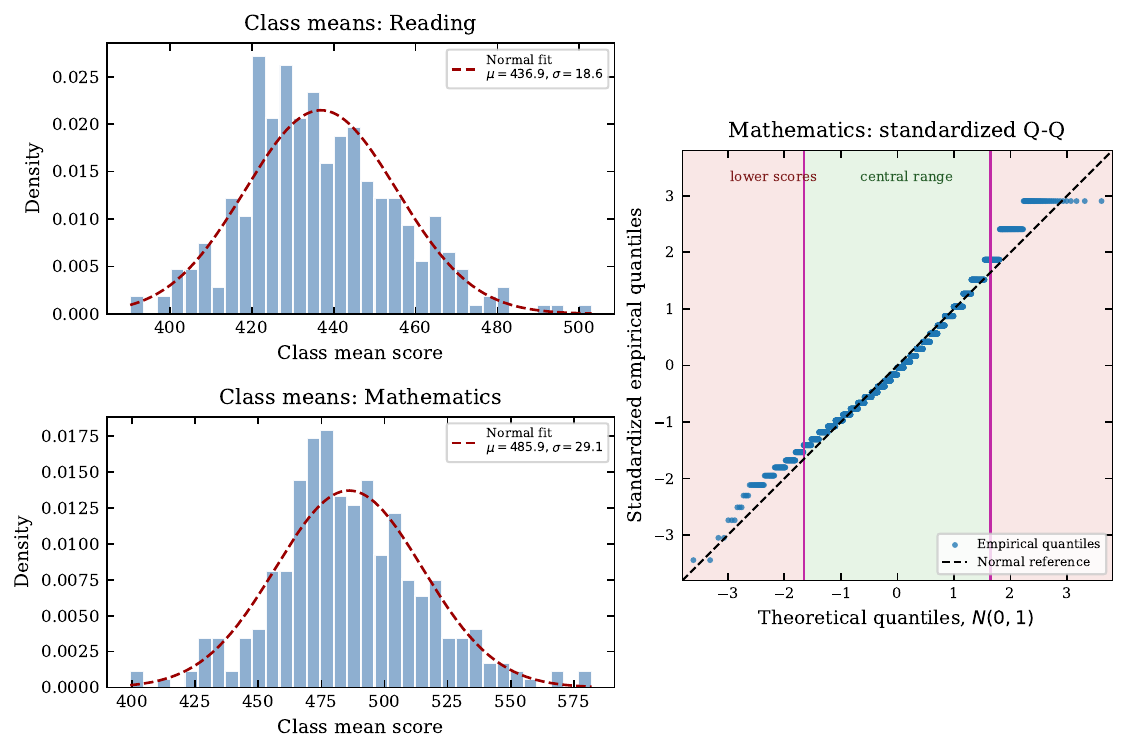}
     \end{minipage}

    \caption{Class-level STAR diagnostics. Left: gender heterogeneity across classes, showing why an additive individual gender coefficient does not represent the full distribution of class composition. Right: reading and mathematics class-level score means, together with a standardized Q-Q diagnostic for Mathematics. In the Q-Q panel, blue points are empirical quantiles and the dashed black line is the Gaussian reference. The green band marks the central range where the Gaussian approximation is most relevant, while the red bands mark lower and upper score extremes. The rightmost extreme is kept as an empirical check rather than fitted separately because it contains only a few repeated score values.}
     \label{fig:gender-heterogeneity}
     \end{figure}
     \subsection{Data and hierarchical representation}
     
     Project STAR randomized kindergarten students within schools to small classes, regular classes, or regular classes with an aide.
The full STAR-and-Beyond public-use dataset is documented by \citet{Achilles2008} and distributed through Harvard Dataverse at \href{https://doi.org/10.7910/DVN/SIWH9f}{https://doi.org/10.7910/DVN/SIWH9f}; it contains the raw student- and school-level records from the longitudinal Tennessee experiment, including demographics, class assignments, school identifiers, teacher information, and achievement outcomes.
After removing missing values, the kindergarten cohort contains 5,745 student records across 322 classes and 79 schools; for the class-as-unit HSCM analysis and diagnostics, the scripts then use a balanced sample of 10 students per class, yielding 3,220 student records across 322 classes.
The main outcome in this paper is Mathematics, written $Y$. Reading, written $B$, is retained as a pre-outcome achievement mechanism because it enters some discovered graph structures and appears in the identified HSCM formulas.
Classes are treated as units and students as sub-units.
The treatment is the student-level small-class assignment, written $A_{ij}=\mathbf 1\{\mathrm{small\ class}\}$, with student covariates given by Gender ($G_{ij}$), Ethnicity ($E_{ij}$), and free-lunch status ($L_{ij}$).
School urbanicity ($S_i$) is treated as an observed class-level covariate, while $U_i$ collects all unobserved unit-level causes, including but not limited to class heterogeneity, latent school context, teacher effects, and other shared classroom factors.

The intervention is not a row-level replacement of $A_{ij}$, but a class-level intervention $\rmdo(Q^a=q_\star^a)$ that fixes the small-class assignment distribution within the class.
In plain terms, the $Q$-quantities used below are class summaries of student-level variables: $Q^a$ is the small-class assignment probability or proportion in a class, $Q^g$ is the gender composition, and Gaussian score mechanisms such as the Mathematics outcome $Q^y$ or the Reading mechanism $Q^b$ are represented by fitted class-specific means and variances.

Figure~\ref{fig:gender-heterogeneity} illustrates the main class-level diagnostics.
The mathematics Q-Q panel compares the standardized empirical score quantiles with a Gaussian reference and highlights the central range separately from the lower and upper extremes.
The upper extreme contains only a few repeated score values, so it is shown as an empirical diagnostic rather than fitted with a separate tail model.

     % \begin{center}
     % \centering
     % \scriptsize
     % \begin{tabular}{ccccccccc}
     % \toprule
     % \texttt{stdntid} & \texttt{gktchid} & $S_i$ & $A_{ij}$ & $G_{ij}$ & $E_{ij}$ & $L_{ij}$ & $Y_{ij}$ & $M_{ij}$ \\
     % \midrule
     % 11311 & 11203801 & 4 & 1 & 1 & 1 & 1 & 413 & 454 \\
     % \bottomrule
     % \end{tabular}
     % \captionof{table}{Example row from the balanced STAR HSCM sample. The row is shown after recoding the raw kindergarten variables into the symbols used by the HSCM analysis.}
     % \label{tab:star-example-row}
     % \end{center}

      \subsection{Baseline regression analysis}
     As baselines, we reproduce standard student-level regression contrasts using the notation introduced in Table~\ref{tab:variables}, following the public STAR regression replication code of \citet{ZCherylSTA207}. Let $A_{ij}$ denote assignment of student $j$ in class $i$ to a small class, let $R_{ij}$ denote assignment to a regular class with an aide, let $Y_{ij}$ be the kindergarten Mathematics outcome, and let $s(i)$ denote the school of class $i$. The OLS specifications 
     $\mathcal{M}^1$-$\mathcal{M}^4$ all report $\hat\beta_A$, the coefficient of the small-class assignment indicator $A_{ij}$.

The pooled OLS contrast is
\begin{align}
  Y_{ij}
  &=
  \alpha
  + \beta_A A_{ij}
  + \beta_R R_{ij}
  + \varepsilon_{ij},
  \label{eq:ols-pooled} \tag{$\mathcal{M}^1$}
\end{align}
the second OLS model adds school fixed effects,
\begin{align}
  Y_{ij}
  &=
  \alpha
  + \beta_A A_{ij}
  + \beta_R R_{ij}
  + \delta_{s(i)}
  + \varepsilon_{ij},
  \label{eq:ols-school} \tag{$\mathcal{M}^2$}
\end{align}
the third one adds observed student controls,
\[
  X^{\mathrm{stu}}_{ij}=(G_{ij},E_{ij},L_{ij}),
\]
\begin{align}
  Y_{ij}
  &=
  \alpha
  + \beta_A A_{ij}
  + \beta_R R_{ij}
  + \gamma_{\mathrm{stu}}^\top X^{\mathrm{stu}}_{ij}
  + \delta_{s(i)}
  + \varepsilon_{ij},
  \label{eq:ols-student} \tag{$\mathcal{M}^3$}
\end{align}
and finally, we consider the full OLS replication. It adds teacher controls,
\[
  X^{\mathrm{full}}_{ij}
  =
  (G_{ij},E_{ij},L_{ij},\text{teacher race},\text{teacher experience},\text{teacher degree}),
\]
and estimates
\begin{align}
  Y_{ij}
  &=
  \alpha
  + \beta_A A_{ij}
  + \beta_R R_{ij}
  + \gamma_{\mathrm{full}}^\top X^{\mathrm{full}}_{ij}
  + \delta_{s(i)}
  + \varepsilon_{ij},
  \label{eq:ols-full} \tag{$\mathcal{M}^4$}
\end{align}
The coefficient $\beta_A$ is therefore a student-level small-class contrast, conditional on the chosen controls.

\begin{table}[t]
    \centering
\caption{Baseline regression estimates for kindergarten mathematics. For M1--M4 and M6, the reported coefficient is $\hat\beta_A$, the coefficient of the small-class assignment indicator $A_{ij}$. For M5, the reported coefficient is the IV class-size coefficient $\hat\theta$.}
\label{tab:ols}
\begin{tabular}{llcccc}
\toprule
Model & Specification & Coef. & SE & $p$-value & $R^2$ \\
\midrule
$\mathcal{M}^1$ & No controls & $\hat\beta_A$=8.096 & 1.597 & $4.0\times 10^{-7}$ & 0.006 \\
$\mathcal{M}^2$ & School fixed effects & $\hat\beta_A$=9.499 & 1.463 & $8.5\times 10^{-11}$ & 0.219 \\
$\mathcal{M}^3$ & Student controls & $\hat\beta_A$=9.380 & 1.415 & $3.4\times 10^{-11}$ & 0.268 \\
$\mathcal{M}^4$ & Full OLS replication & $\hat\beta_A$=9.432 & 1.419 & $3.0\times 10^{-11}$ & 0.269 \\
$\mathcal{M}^5$ & IV class-size contrast & $\hat\theta$=-1.219 & 0.163 & $7.0\times 10^{-14}$ & -- \\
$\mathcal{M}^6$ & Random class intercept & $\hat\beta_A$=9.099 & 2.263 & $5.8\times10^{-5}$ & -- \\
\bottomrule
\end{tabular}
\end{table}

The instrumental-variable replication, denoted $\mathcal{M}^5$, uses assignment indicators as instruments for actual class size $C_{ij}$. It decomposes into two steps:
\begin{align}
  C_{ij}
  &=
  \pi_0
  + \pi_A A_{ij}
  + \pi_R R_{ij}
  + \Pi^\top X_{ij}
  + \delta_{s(i)}
  + \nu_{ij}, \tag{$\mathcal{M}^5_1$}
  \label{eq:first-stage}\\
  Y_{ij}
  &=
  \alpha
  + \theta \widehat C_{ij}
  + \gamma^\top X_{ij}
  + \delta_{s(i)}
  + \varepsilon_{ij}. \tag{$\mathcal{M}^5_2$}
  \label{eq:second-stage}
\end{align}
We get then a class-size contrast estimate $\hat{\theta}$ rather than another estimate of $\beta_A$.

The corresponding non-causal hierarchical regression baseline, denoted $\mathcal{M}^6$, can be written as a random-intercept or Bayesian OLS model:
\begin{align}
  Y_{ij}
  &=
  \alpha
  + \beta_A A_{ij}
  + \beta_R R_{ij}
  + \gamma^\top X_{ij}
  + b_{\mathrm{class}(i)}
  + \varepsilon_{ij},
  \qquad
  b_c \sim \mathcal{N}(0,\sigma_b^2). \tag{$\mathcal{M}^6$}
  \label{eq:hierarchical-baseline}
\end{align}

We fit M6 with the same fixed controls as M4 and a random intercept at the class level. The resulting small-class coefficient is $\hat\beta_A=9.099$ (SE $2.263$), so the hierarchical regression baseline remains close to the OLS small-class contrasts but with larger uncertainty.

This model is valuable because it acknowledges between-class heterogeneity, but its coefficient $\beta_A$ remains a regression coefficient with a random class intercept; it treats heterogeneity as a residual random effect rather than as a distributional causal mechanism.
%In other words, the model can say that some classes have higher or lower baseline mathematics scores, and OLS can adjust for observed individual covariates such as gender or free lunch, but neither model represents the class as an object whose internal composition changes the intervention itself.

     Results are given in Table \ref{tab:ols}.
The models which are estimating $\hat\beta_A$ ($\mathcal{M}^i$ for $i\in\{1,2,3,4,6\}$) are providing similar results, but we trust more the value of $\mathcal{M}^6$ which is benefiting of the hierarchical structure of the data. 
The IV model $\mathcal{M}^5$ is giving another information, with a class-size contrast estimate. The sign of the estimator, negative, is expected.

\begin{figure}[b]
\centering
%\noindent\begin{minipage}[t]{0.7\textwidth}
\vspace{0pt}
\centering
\includegraphics[height=3.8cm, trim = 0 15cm 4cm 1.3cm, clip=TRUE]{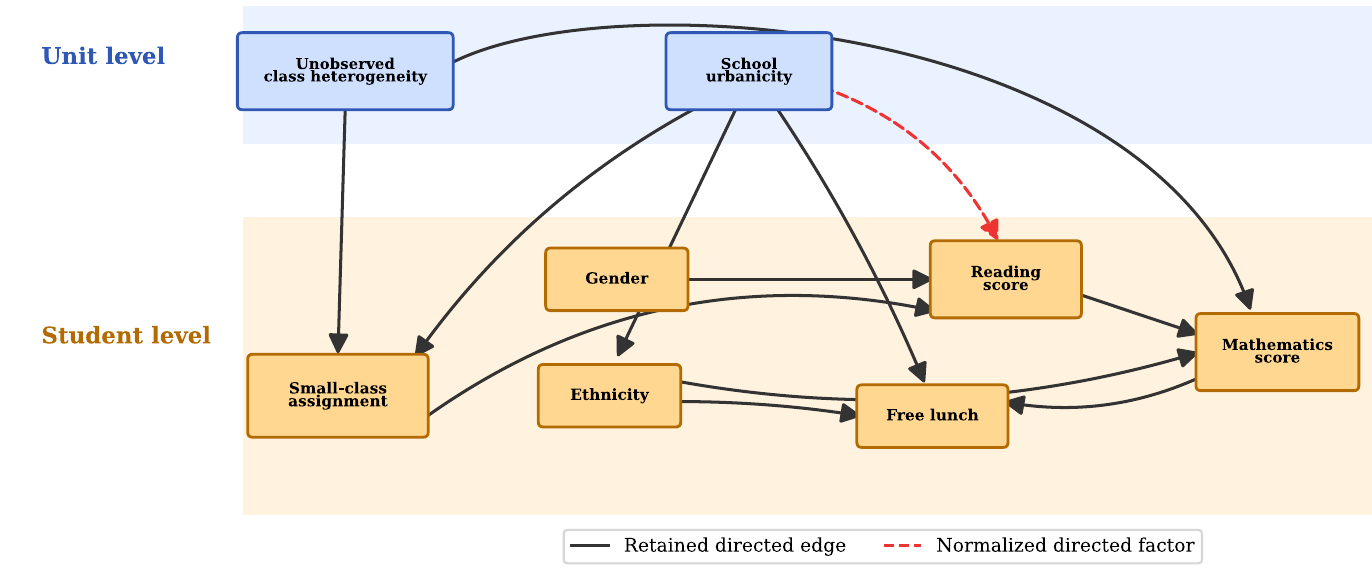}
\includegraphics[height=3.8cm, trim = 0 0.5cm 0cm 13.7cm, clip=TRUE]{Fig/ExactBIC_merged_non_q_and_q_directed.pdf}
%\end{minipage}%
% \hspace{0.02\textwidth}%
% \begin{minipage}[t]{0.26\textwidth}
% \vspace{0pt}
% \centering
% \scriptsize
% \captionof{table}{STAR variables and HSCM objects used in the ExactBIC graph.}
% \label{tab:exactbic-variable-map}
% \begin{tabular}{lll}
% \toprule
%  Symbol & HSCM object \\
% \midrule
%  $A$ & $Q^a$ small-class \\
%  $B$ & $Q^{b\mid a,g}$ reading \\
%  $Y$ & $Q^{y\mid e,b}$ mathematics \\
%  $G$ & $Q^g$ gender \\
%  $E$ & $Q^e$ ethnicity \\
% % $L$ & $Q^l$ lunch \\
%  $S$ & unit context, school urbanicity \\
%  $U$ & class heterogeneity \\
% \bottomrule
% \end{tabular}
% \end{minipage}
\caption{ExactBIC-derived graphs for the STAR HSCM analysis. The left panel shows the directed ExactBIC graph over the flat STAR variables, after adding the unit-level context used by the HSCM analysis. The right panel shows the corresponding collapsed and augmented $Q$-variable graph for the Mathematics outcome. Blue nodes are unit-level context variables, orange nodes are observed variables or fitted $Q$-mechanisms, and the green node is the target class-level Mathematics outcome distribution $Q^y$. Solid arrows are retained directed dependencies; dashed red arrows mark factors normalized in the stabilized numerical evaluation.}
\label{fig:exactbic-effective-graph}

\end{figure}

However, in the previous standard baselines, the heterogeneity is absorbed into fixed effects, controls, or random intercepts; it is not decomposed into class-level $Q$-mechanisms such as the treatment distribution $Q^a$, gender composition $Q^g$, or response factors depending jointly on treatment and composition.
%This is precisely where a causal HSCM analysis becomes necessary.
For STAR, the scientific question is not only whether an individual assigned to a small class has a different expected score, but how a class-level intervention on the distribution of small-class assignment propagates through heterogeneous classroom mechanisms.
%OLS, IV, and hierarchical Bayesian/OLS baselines therefore include gender and class variation descriptively, but they do not treat class composition or treatment-by-gender mechanisms as intervention-level distributions.
This distinction motivates the use of HSCM  below.

\subsection{Causal graph discovery and associated HSCM}

The STAR HSCM analysis starts from graph structures learned on the flat discovery variables $A, B, Y, G, E, L$ and $S$, since no causal graph is assumed a priori. We compared PC \citep{pc}, FCI \citep{fci}, DirectLiNGAM \citep{directlingam}, and ExactBIC \citep{exactbic}. The graph reported in Figure~\ref{fig:exactbic-effective-graph} (left) is the ExactBIC graph selected for the HSCM analysis: it is not claimed to be the unique recovered causal graph, but it was the most informative discovered structure for the STAR translation, because treatment, school context, class composition, reading, and mathematics remained connected after collapse and augmentation.

For the ExactBIC run, the flat graph is first read as a directed graph over the observed STAR variables. The HSCM construction then adds the latent class-level context used in the analysis, in particular the unobserved class heterogeneity node $U$ and the school-context node $S$. Since the outcome of interest is kindergarten Mathematics, the graph is then collapsed and augmented for the class-level Mathematics outcome mechanism $Q^y$. This gives the $Q$-variable graph shown in the right panel of Figure~\ref{fig:exactbic-effective-graph}.

     \begin{table}[t]
     \centering
     \caption{Class-level $Q$-variables used by the STAR HSCM evaluator. These are distributional summaries or mechanisms fitted at the class level from the student-level variables in Table~\ref{tab:variables}.}
     \label{tab:qvariables}
     \begin{tabular}{lll}
     \toprule
     Symbol & Type & Interpretation \\
     \midrule
     $Q^a$ & Bernoulli/proportion & Small-class assignment distribution within a class \\
     $Q^g$ & Bernoulli/proportion & Gender composition within a class \\
     $Q^e$ & Bernoulli/proportion & Ethnicity composition within a class \\
     $Q^l$ & Bernoulli/proportion & Free-lunch composition within a class \\
    $Q^b$ & Gaussian score mechanism & Reading score distribution for the class \\
$Q^y$ & Gaussian score mechanism & Mathematics score distribution for the class \\
$Q^{b\g a,g}$ & Conditional mechanism & Reading mechanism conditional on treatment and gender composition \\
$Q^{y\g e,b}$ & Conditional mechanism & Mathematics mechanism conditional on ethnicity and reading mechanisms \\
     \bottomrule
     \end{tabular}
     \end{table}

% \begin{figure}[t]
% \centering
% \includegraphics[width=0.98\textwidth, trim = 0 0 0 1.5cm, clip=TRUE]{Fig/ExactBIC_with_factor1_effective_graph_curved.pdf}
% \caption{ExactBIC-derived graph for the STAR HSCM analysis. The graph is learned on the flat discovery variables and then interpreted after the HSCM collapse/augmentation step. Blue nodes represent unit-level context variables, orange nodes represent collapsed or augmented $Q$-mechanisms, and the green node is the target class-level Mathematics outcome distribution $Q^y$. Solid arrows denote directed dependencies retained from the ExactBIC graph and its HSCM translation, while the dashed red arrow marks the Reading factor $P(Q^{b\g a,g}\g S)$ normalized in the stabilized numerical evaluation.}
% \label{fig:exactbic-effective-graph-Qvariables}
% \end{figure}

The STAR HSCM analysis starts from graph structures learned on the flat discovery variables $A, B, Y, G, E, L$ and $S$, as there is no a priori knowledge on the causal graph. We compared four discovery procedures on this flat representation: PC \citep{pc}, FCI \citep{fci}, DirectLiNGAM \citep{directlingam}, and ExactBIC \citep{exactbic}. The graph reported in Figure \ref{fig:exactbic-effective-graph} is the ExactBIC graph, not because it is claimed to be the uniquely recovered causal graph, but because among the discovered candidates it gave the most informative structure for the HSCM translation: the treatment, achievement, composition, and school-context variables remained connected in a way that produced a non-degenerate class-level functional after collapse and augmentation.
For the ExactBIC run, we used one score-based DAG stored in the discovery output, with the school-urbanicity/ethnicity relation oriented as Urbanicity ($S$) $\to$ Ethnicity ($E$) for substantive interpretability. This is one plausible graph, not a claim that the discovery step has recovered a unique causal structure.

Before identification, the HSCM construction also adds the latent class-level confounding edges Unobserved class heterogeneity ($U$) $\to$ Small-class assignment ($A$) and Unobserved class heterogeneity ($U$) $\to$ Mathematics outcome ($Y$), because the target outcome in this paper is Mathematics.
The resulting hierarchical graph is then collapsed and augmented for $Q^y$ (the class-level Mathematics outcome mechanism), giving the $Q$-variable graph in Figure~\ref{fig:exactbic-effective-graph}.
The new variables, introduced in the collapsed and augmented graph, are summarized in Table \ref{tab:qvariables}.

     \subsection{Identified formulae}
     
In the second step, the transformed graph is passed to \texttt{pyAgrum} to identify the query. 
It gives the following functional after the translated graph has been collapsed and augmented for the mathematics outcome:
\begin{align}
\tau_{\mathrm{BIC}}(q^a_\star)
&=
\sum_{q^e,q^g,q^{y\g e,b},q^{b\g a,g},s}
\pr(q^e\g s)
\pr(q^{b\g a,g}\g s)
\pr(q^{y\g e,b})
\Exp{Q^y \g Q^a=q^a_\star,Q^e,Q^g,Q^{y\g e,b},Q^{b\g a,g}}
\pr(q^g)\pr(s).
\label{eq:bicstar}
\end{align}
In this expression, $\pr(q^{y\g e,b})$ is the class-level Mathematics outcome distribution induced by the discovered Reading ($B$) $\to$ Mathematics outcome ($Y$) and Ethnicity ($E$) $\to$ Mathematics outcome ($Y$) parents, $\pr(q^e\g s)$ encodes the substantively oriented Urbanicity ($S$) $\to$ Ethnicity ($E$) relation, and $\pr(q^{b\g a,g}\g s)$ comes from the Reading ($B$) mechanism with Small-class assignment ($A$), Gender ($G$), and Urbanicity ($S$) in the translated graph.
The expectation term is evaluated twice, at $Q^a=1$ and $Q^a=0$, and the remaining $Q$-variables are marginalized by the product of fitted density factors.
This is a do-calculus functional in the sense of \citet{Pearl2009} and \citet{weinstein2023HSCM}; its product weights are also close in spirit to importance-sampling and inverse-probability weighting constructions \citep{RubinsteinKroese2016,HorvitzThompson1952,RobinsHernanBrumback2000}.
The formula therefore makes clear why multi-parent $Q$-estimation matters: the identified effect is a product of conditional densities and outcome-response factors, not a single regression coefficient.

\subsection{Estimation}

Finally, in the third step, this query is estimated with the fitted model families used in the run: Bernoulli models for Small-class assignment ($A$), Gender ($G$), Ethnicity ($E$), and free lunch ($L$), a categorical model for Urbanicity ($S$), and two-component Gaussian mixtures for Reading ($B$) and Mathematics outcome ($Y$).
After this translation, the class-as-unit HSCM model gives the estimates reported in Table~\ref{tab:HSCMstar}, together with the separate normalized-factor diagnostic for the Reading mechanism.

          \begin{table}[t]
\centering
\small
\caption{STAR HSCM estimates and normalized-factor diagnostic. The Mathematics row is the main outcome analysis, where the outcome is denoted $Y$; the Reading row is a separate diagnostic run used to inspect the normalized $P(Q^{b\g a,g}\g S)$ factor.}
\label{tab:HSCMstar}
\begin{tabular}{llccc}
\toprule
Run & Target & $E[\cdot | \rmdo(1)]$ & $E[\cdot | \rmdo(0)]$ & ATE \\
\midrule
ExactBIC & Mathematics outcome ($Y$) & 2402.75 & 2365.95 & 36.80 \\
ExactBIC, normalized factors & Reading ($B$) diagnostic & 1788.10 & 1766.47 & 21.63 \\
\bottomrule
\end{tabular}
\end{table}

     \begin{figure}[b]
     \centering
     \includegraphics[width=0.98\textwidth, trim = 0 0 0 1.3cm, clip=TRUE]{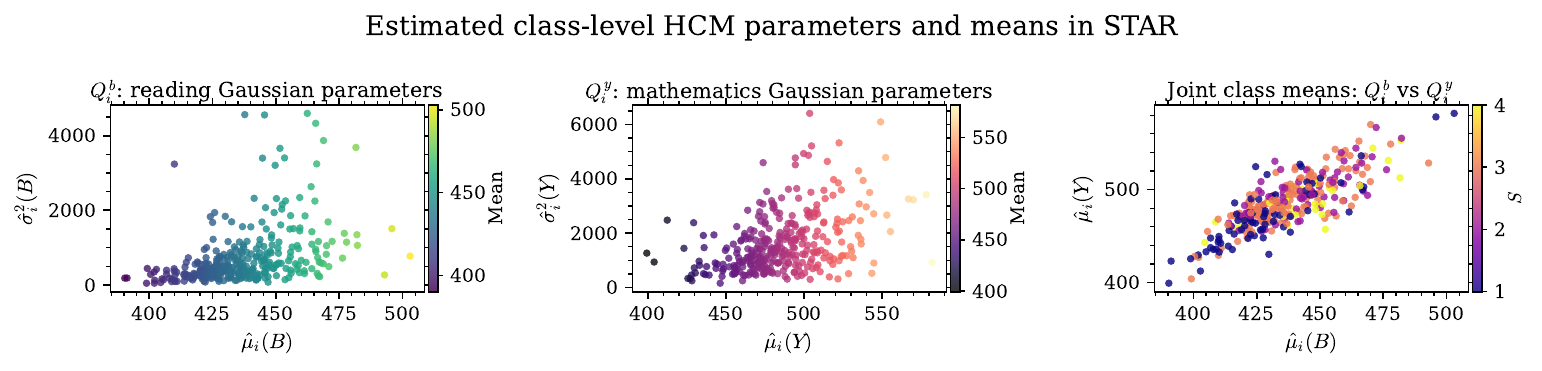}
    \caption{Estimated class-level HSCM response parameters and means in STAR\@. Reading ($Q^b$) and the Mathematics outcome ($Q^y$) are represented by Gaussian summaries $(\hat\mu_i,\hat\sigma_i^2)$, and the right panel reports the joint class means used to inspect the relationship between the two score mechanisms. In the right panel, colors encode the class-level urbanicity category $S$; the panel is a descriptive diagnostic rather than an oriented causal graph.}
     \label{fig:q-parameters-means}
     \end{figure}

%\clearpage
     These values should not be read as direct replacements for the student-level OLS coefficient, but the comparison is still essential.
     Across the three OLS specifications with school fixed effects or student controls, the small-class coefficient is approximately $9$ to $9.5$ mathematics points, whereas the ExactBIC HSCM mathematics contrast in Table~\ref{tab:HSCMstar} is $36.80$ points.
     OLS gives a student-level contrast for a binary treatment indicator, conditional on observed covariates and school fixed effects; IV gives a class-size contrast using assignment as an instrument.
     The HSCM estimand is instead a class-level intervention on $Q^a$, with additional dependence through class-specific distributions.
     The fact that the ExactBIC HSCM contrast remains much larger than the OLS coefficient is a warning sign: once the hierarchy is ignored, the target no longer contains the class-composition mechanisms that drive the HSCM functional.
     The HSCM magnitude therefore describes the behaviour of the translated hierarchical estimand, while the separate normalized-factor diagnostic below shows that the numerical result is sensitive to the scaling of multiplicative factors retained in the identified formula.
     
     We also report the speed up gain with our parallel implementation with a CPU. In sequential, the evaluation take 3.11 seconds, while with 4 processes it took 1.9 seconds, with a speed up of 1.64.

     Figure~\ref{fig:q-parameters-means} shows that the response mechanisms are not constant across classes.
     For both Reading and Mathematics, classes with larger fitted means also tend to have larger fitted variances, so the Gaussian $Q$-summaries capture both level differences and heteroskedasticity across classrooms.
     This matters for the HSCM numerical estimator because an intervention on the class-level treatment distribution is evaluated through these class-specific response mechanisms, not through a single pooled outcome regression.
     The right panel shows a strong positive association between class-level Reading and Mathematics means, indicating that the two achievement mechanisms share substantial classroom-level structure.
     Consequently, the mathematics effect should not be interpreted as an isolated student-level coefficient: it is evaluated in a distribution of classrooms whose baseline achievement and score variability differ markedly.

%      \begin{table}[!htbp]
% \centering
% \small
% \caption{STAR two-intervention HSCM CPU evaluation runtime. Times are in seconds. The row reports the ExactBIC STAR mathematics speed test and the corresponding two-intervention ATE; it is kept as the real-data CPU reference, while the GPU experiment in Figure~\ref{fig:gpu-speedup-cuda} uses larger synthetic HSCM workloads to measure tensor batching.}
% \label{tab:parallel-runtime}
% \begin{tabular}{lcccc}
% \toprule
% Workload & Sequential & 4 processes & Best speedup & ATE \\
% \midrule
% ExactBIC STAR HSCM evaluation & 3.11 & 1.90 & $1.64\times$ & 36.80 \\
% \bottomrule
% \end{tabular}
% \end{table}

     \subsection{Effective graphs and factor normalization}
     
     The normalized-factor diagnostic in Table~\ref{tab:HSCMstar} makes explicit the difference between a nominal identified functional and the stabilized functional actually evaluated by the numerical code.
     This diagnostic is needed because the effective sample size for HSCM estimation is the number of classes, not the number of student rows.
     In the balanced STAR analysis there are 322 classes and only 10 sampled students per class, so class-level mechanisms with several parents are estimated with substantial uncertainty.
     When such uncertain mechanisms enter a formula as multiplicative probability terms, a single poorly estimated term can dominate the final numerical product.
     For ExactBIC, the diagnostic run concerns the Reading ($Y$) mechanism and the normalized term was $P(Q^{y\g a,g}\g S)$, which is the distribution of the class-level Reading mechanism conditional on Small-class assignment ($A$) and Gender ($G$), further conditioned by Urbanicity ($S$).
     Normalizing such a term does not remove variables from the nominal causal graph and should not be read as a new estimand selected after seeing the answer.
     It is a stability analysis: it asks whether the qualitative contrast is being carried by the response mechanism itself or by the raw scale of one uncertain fitted probability term.
     The effective graph in Figure~\ref{fig:exactbic-effective-graph} visualizes this distinction: dashed red arrows are dependencies present in the nominal graph whose factor contribution is normalized in the stabilized functional.
     
     The ExactBIC gender factor analysis clarifies what this means numerically.
     In Table~\ref{tab:HSCMstar}, the normalized factor $P(Q^{y\g a,g}\g S)$ is stabilized before the two intervention evaluations, so its raw scale does not dominate the product. The corresponding factor-level traces vary on roughly $10^3$ scales in the raw product, which is precisely the numerical symptom that motivates the normalized analysis.
     It nevertheless marks an important part of the estimand: the graph says that the class-level Reading ($Y$) response should depend on Small-class assignment ($A$), Gender ($G$) composition, and Urbanicity ($S$).
     When the same run is analysed in unit context, the only non-zero factor change comes from the fitted response term $P(Q^y\g Q^a,Q^g,Q^{y\g a,g})$, namely Reading ($Y$) conditional on Small-class assignment ($A$), Gender ($G$), and the class-level Reading mechanism $Q^{y\g a,g}$, whose mean changes from $434.83$ under $\rmdo(Q^a=0)$ to $440.15$ under $\rmdo(Q^a=1)$.
     The resulting delta, $5.32$, explains why the gender mechanism matters even when the conditional factor is normalized for stability.

     \begin{figure}[t]
     \centering
     \includegraphics[width=0.88\textwidth, trim = 0 0 0 0.8cm, clip=TRUE]{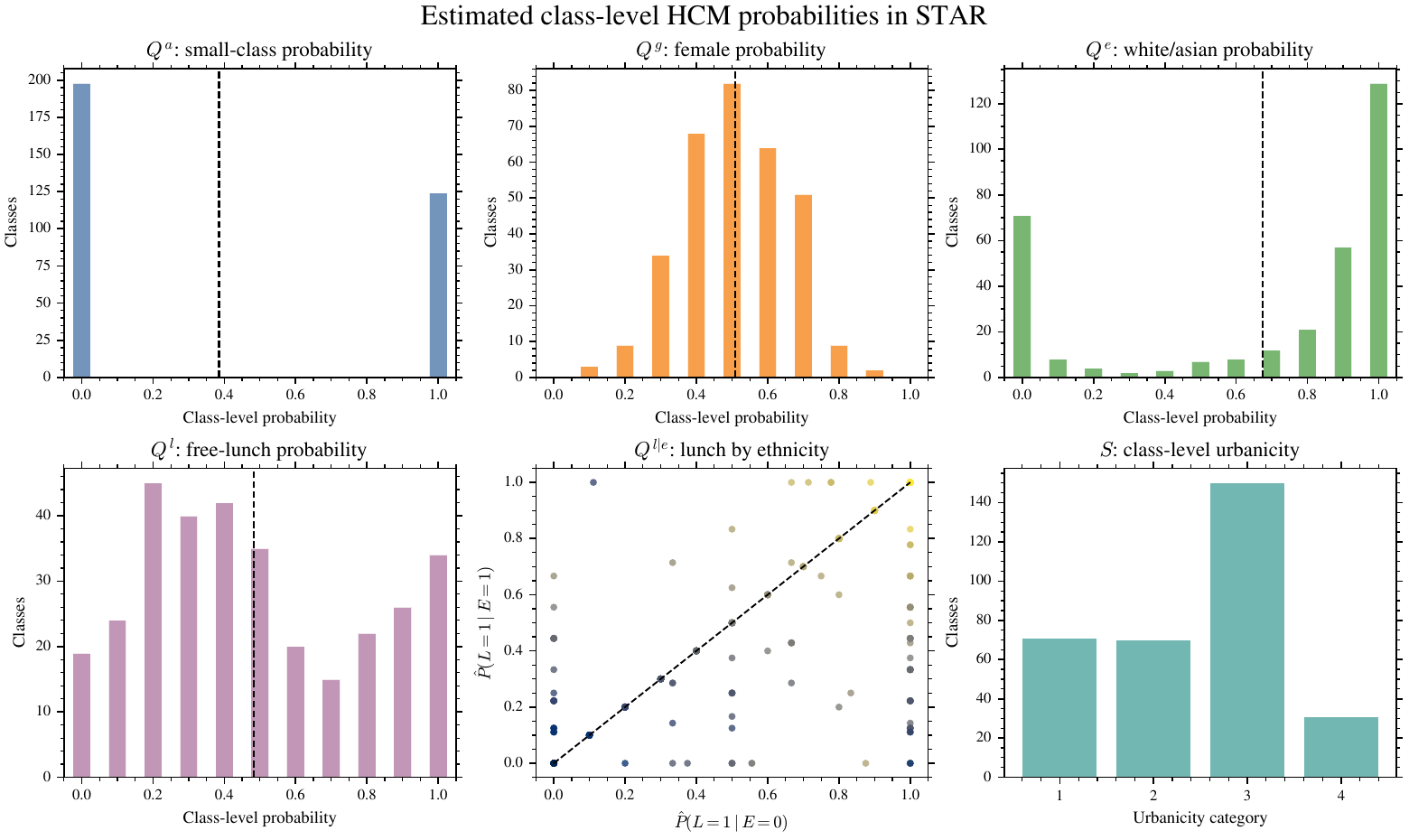}
    \caption{Estimated class-level HSCM probability quantities used in the STAR analysis. Binary sub-unit variables are represented by class-level Bernoulli probabilities for Small-class assignment ($Q^a$), Gender ($Q^g$), Ethnicity ($Q^e$), and free lunch ($Q^l$); $Q^{l\g e}$ summarizes the free-lunch mechanism conditional on Ethnicity; and $S$ is the observed class-level Urbanicity variable. In the conditional free-lunch scatter, point colors encode the urbanicity category $S$; in the $S$ panel itself, the bar color is only a visual grouping choice.}
     \label{fig:q-probabilities}
     \end{figure}
     
     \subsection{Diagnostics and interpretation}

     The STAR analysis also illustrates why diagnostics are essential.
     The experiments expose a central operational issue: an identified formula may contain multi-parent $Q$-variables whose density factors are numerically unstable.
     In a small class-level dataset, these fitted terms can carry large estimation uncertainty; the normalized-factor runs rescale them to check whether the final contrast is stable to this raw product scale.
     In the ExactBIC STAR run, the normalized factor was the gender-dependent term $P(Q^{y\g a,g}\g S)$.
     The diagnostic is scientifically useful only when it is read at the factor level: factor levels show which terms dominate the product numerically, while unit-context factor changes show which term changes between the two interventions.
     
     The gender-composition diagnostic in Figure~\ref{fig:gender-heterogeneity} shows that class composition varies substantially across classes; an OLS coefficient on \texttt{female} adjusts for an average individual difference, but it does not represent the distribution of class composition.
     HSCM explicitly represents such within-class distributions through the estimated probability $Q$-quantities in Figure~\ref{fig:q-probabilities} and the response parameters and means in Figure~\ref{fig:q-parameters-means}.
     The ExactBIC factor analysis above shows how gender composition enters the intervention path through $Q^{y\g a,g}$.
     The current implementation remains two-level and treats schools only through unit-level covariates rather than as a full student--class--school hierarchy.
     This is the main substantive lesson of the STAR analysis.
     Earlier non-causal hierarchical or graphical summaries could describe that classes differ, but they did not make class heterogeneity part of an identified intervention functional.
     By combining class-level $Q$-heterogeneity with factor analysis of the identified formula, the HSCM pipeline shows that part of the small-class effect can be hidden when class composition and multi-parent distributional factors are treated only as background variation.
     In that sense, the previous hierarchical graph analysis tended to smooth or understate the intervention contrast, whereas the causal HSCM analysis gives a new interpretation of STAR: the effect of class size is not only an average student-level coefficient, but a distributional class-level effect whose magnitude depends on which class mechanisms are included, estimated, or normalized in the numerical functional.
    
     The substantive conclusion is therefore conservative.
     HSCM does not simply ``improve'' OLS by producing a larger number; it changes the estimand from a student-level coefficient to a distributional class-level intervention.
     In STAR this change is scientifically meaningful, but it also makes graph choice, positivity, and completeness of the fitted local models part of the reported result rather than secondary implementation details.

     %% =============================================================================
     \section{Conclusion}
     \label{sec:conclusion}
     %% =============================================================================
     
     We developed a scalable automatic identification and estimation implementation for Hierarchical Structural Causal inference.
     The pipeline starts from a hierarchical graph, applies the HSCM transformations needed for identification, obtains a symbolic do-calculus estimand from \texttt{pyAgrum}, adapts it into closed-form HSCM formulas, and evaluates the associated AST using fitted local probability and response models.
     This turns HSCM from a formal identification framework into an operational workflow for nested data.
     The key computational point is that the AST does not leave the formula as an opaque expression: it decomposes the estimand into many local conditional densities, response models, and marginalization branches that can be estimated independently and then recombined into a single causal functional.
     The implementation is available at \href{https://github.com/AI-vidence/hierarchicalcausalmodels}{\texttt{AI-vidence/hierarchicalcausalmodels}}, so the graph transformations, identification calls, estimators, diagnostics, and STAR replication artefacts are reproducible. 
     
     The experiments validate the pipeline on known structures and reveal the practical regimes in which estimation is stable.
     The STAR application demonstrates the methodological payoff: when the intervention is assigned at the class level and outcomes are measured at the student level, the natural estimand is hierarchical.
     OLS, IV, and non-causal hierarchical summaries remain essential benchmarks, but they do not encode interventions on within-class treatment distributions or class-composition mechanisms.
     The same framework also clarifies why large hierarchical applications are computationally feasible: adding units, factors, or sampled branches increases the number of local estimation tasks, but it does not require redefining the estimand or replacing it by a flat pooled regression.
     
     The remaining limitations point directly to the next statistical developments.
     The present implementation is two-level, whereas STAR is naturally a student--class--school system; extending scalable HSCM estimation to deeper hierarchies is therefore necessary for a fully faithful education application.
     Multi-parent $Q$-density estimation is the main statistical bottleneck, because missing or weakly estimated factors change the effective estimand rather than merely adding numerical noise.
     Uncertainty quantification for normalized or otherwise stabilized functionals also remains open.
     future work should therefore develop robust multi-parent estimators, extend the implementation to deeper hierarchies, preserve the parallel structure of the AST evaluator, and connect the numerical backend to symbolic DSLs such as \texttt{y0} so that formula manipulation and estimation can share a common representation. For discovery-driven applications, the next step is to evaluate families of plausible DAGs rather than a single graph, using cluster-DAG ideas to group graphs that agree on coherent blocks of variables and to report how the HSCM estimand changes across those graph clusters.
     
     %% =============================================================================
     \section*{Author contributions}
     All authors contributed to the methodology.
     A.E., Y.O., S.P., A.E.O., A.D. and J.A. contributed to the software devlopment and experiments running. 
     All authors contributed to the analysis of the results. 
     D.C., J.A, E.D. and M.C. contributed to the writing.

     \section*{Data availability}
     The full Project STAR public-use data are available as the STAR-and-Beyond dataset documented by \citet{Achilles2008} and hosted by Harvard Dataverse at \href{https://doi.org/10.7910/DVN/SIWH9f}{this link}. Code and generated artefacts are maintained at \href{https://github.com/AI-vidence/hierarchicalcausalmodels}{github.com/AI-vidence/hierarchicalcausalmodels}.

     \end{document}